\documentclass[journal]{IEEEtran}
\ifCLASSINFOpdf
\else
\fi
\usepackage{times}  
\usepackage{helvet} 
\usepackage{courier}  
\usepackage[hyphens]{url}  
\usepackage{graphicx} 
\usepackage{caption} 
\usepackage{stfloats}
\usepackage{latexsym}
\usepackage{array}
\newcolumntype{C}[1]{>{\centering\arraybackslash}p{#1}}
\usepackage{algorithm}
\usepackage{algorithmic}   
\usepackage{amsmath}
\usepackage{amsfonts}
\usepackage{url}
\usepackage{float}
\usepackage[switch]{lineno}
\usepackage{booktabs}
\usepackage{tabularx}
\usepackage{threeparttable}  
\usepackage{graphicx}
\usepackage{subfigure}
\usepackage{multirow}
\usepackage{amssymb}
\usepackage{placeins}
\usepackage{color}
\usepackage{verbatim}
\usepackage{bm}
\usepackage{hyperref}
\usepackage{adjustbox}
\begin{document}
%
\title{High-Dimensional Noise to Low-Dimensional Manifolds: A Manifold-Space Diffusion Framework for Degraded Hyperspectral Image Classification}
%
%
%

\author{Boxiang~Yang,
	    Ning~Chen,
        Xia Yue,
        Yichang Luo,
        Yingbo Fan,
        Haoyuan Zhang,
        Haoyu Ma,
        Jun Yue,
        and Shanjun Mao
	\thanks{This work was supported in part by the National Natural Science Foundation of China under Grant 62505008 \emph{(Corresponding author:Ning Chen.)}}
	\thanks{Boxiang Yang, Ning Chen, Haoyu Ma and Shanjun Mao are with the Institute of Remote Sensing and Geographic Information System, Peking University, Beijing 100871, China, with the Beijing Key Laboratory of Spatio-temporal Perception and Urban Resilience, Peking University, Beijing 100871, China (e-mail: yangboxiang@stu.pku.edu.cn, chenning0115@pku.edu.cn, xhaoyu491@gmail.com, sjmao\_pku@163.com).}

     \thanks{Xia Yue is with School of Computer Science and Engineering, Central South University, Changsha 410083, China (e-mail: yuexia486@gmail.com).}

     \thanks{Yichang Luo is with the Aerospace Information Research Institute, Chinese Academy of Sciences, Beijing, China (e-mail: luoyichang24@mails.ucas.ac.cn).}
    
    \thanks{Yingbo Fan is with the Institute of energy, Peking University, Beijing, China (e-mail: ybfan@stu.pku.edu.cn).}

    \thanks{Haoyuan Zhang is with the School of Mechanics and Engineering Science, Peking University, Beijing, China (email: zhanghaoyuan@pku.edu.cn).}
    
    \thanks{Jun Yue is with the School of Automation, Central South University, Changsha 410083, China (e-mail: junyue@csu.edu.cn).}

    }

%
%

\markboth{}%
{Shell \MakeLowercase{\textit{et al.}}: Bare Demo of IEEEtran.cls for IEEE Journals}
%



\maketitle

\begin{abstract}
Recently, Hyperspectral Image (HSI) classification has attracted increasing attention in remote sensing. However, HSI data are inherently high-dimensional but low-rank, with discriminative information concentrated on a low-dimensional latent manifold. In real-world remote sensing scenarios, the superposition of multiple degradation factors disrupts this intrinsic manifold structure, driving samples away from their original low-dimensional distribution and introducing substantial redundant and non-discriminative variations. To better handle this challenge, this paper proposes a manifold-space diffusion framework (MSDiff) for robust hyperspectral classification under complex degradation conditions. Specifically, the proposed method first maps high-dimensional, degradation-affected HSI data into a compact low-dimensional manifold through a discriminative spectral–spatial reconstruction task, preserving class semantics and reducing redundant variations. A diffusion-based generative model is then applied to regularize the spectral–spatial distribution within the manifold, enabling progressive refinement and stabilization of latent features against residual degradations. The key advantage of the proposed framework lies in performing diffusion-based distribution modeling directly on the low-dimensional manifold, effectively decoupling degradation-induced disturbances from intrinsic discriminative structures and enhancing representation stability under complex degradations. Experimental results on multiple hyperspectral benchmarks demonstrate consistent performance improvements over state-of-the-art methods under diverse composite degradation settings. The code will be available at https://github.com/yangboxiang1207/MSDiff.
\end{abstract}


%
\IEEEpeerreviewmaketitle

\section{INTRODUCTION}
%
%
%
%

    \IEEEPARstart{H}{yperspectral} Imaging, as an important branch of remote sensing technology, enables the acquisition of fine-grained surface reflectance characteristics across continuous and narrow spectral bands, thereby providing substantially higher spectral resolution and richer information content than traditional multispectral imaging  \cite{10815625}. Owing to its distinctive capability in characterizing complex surface compositions and subtle spectral differences, HSI has been widely applied in a variety of fields, including precision agriculture  \cite{RAM2024109037}, mineral exploration \cite{ADEP2017106}, urban planning \cite{DELUCA2024112}, ecological and environmental monitoring \cite{11322757}, and disaster assessment \cite{11045957}, emerging as a key technology for intelligent Earth observation systems. With the improvement of sensor accuracy and acquisition efficiency, HSI data scale and application scenarios have expanded rapidly, making HSI-based object classification and semantic understanding core components of intelligent remote sensing analysis \cite{SONG2025103285}. However, practical HSI is inevitably affected by superimposed composite degradations \cite{9552462}, which not only reduce signal-to-noise ratio but also introduce nonlinear spectral-spatial heterogeneous perturbations, resulting in complex unstable data distributions that pose significant challenges for practical HSI classification \cite{10637422}.

\begin{figure}[!h]
	\centering 
	\includegraphics[width=0.45\textwidth]{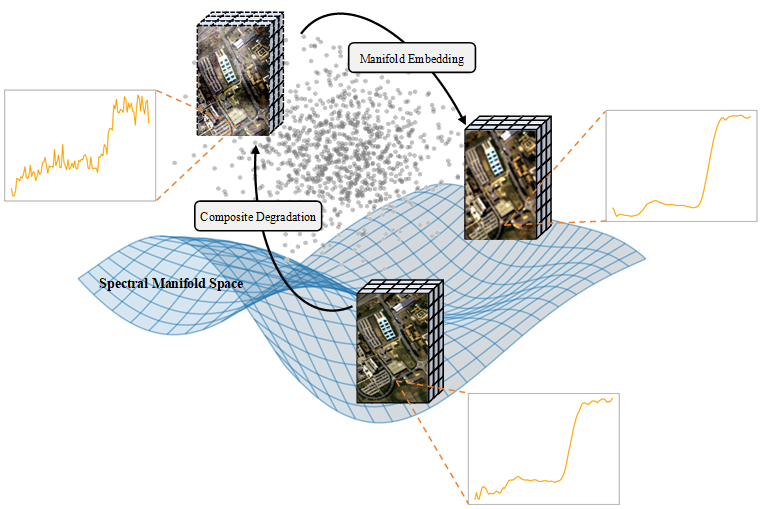} 
	\caption{Spectral manifold hypothesis under composite degradation, illustrating how composite degradations drive hyperspectral observations away from an intrinsic low-dimensional spectral manifold into a more dispersed high-dimensional space.} 
	\label{fig:manifold} 
\end{figure}

Although HSI contain a large amount of spectral information, their effective discriminative content is often highly redundant \cite{10645292} and concentrated within a latent subspace whose dimensionality is far lower than that of the original observation space. As a result, hyperspectral data are commonly regarded as typical high-dimensional but low-rank data \cite{8359412,8603806}, whose intrinsic structure is more appropriately characterized using low-dimensional latent representations or manifold-based models \cite{8677267}. However, the difficulty of HSI classification arises not only from the high dimensionality of the data, but more fundamentally from the strong sensitivity of its intrinsic structure to external perturbations. Under practical remote sensing conditions, the superposition of multiple degradation factors can disrupt the original low-dimensional structure of hyperspectral data, causing sample distributions to deviate from their underlying manifold and expand into a higher-dimensional observation space, thereby introducing a large amount of redundant and non-discriminative information \cite{10890865}. This phenomenon is conceptually illustrated in Fig.~\ref{fig:manifold}, where composite degradations drive hyperspectral observations away from a compact spectral manifold into a more dispersed high-dimensional space. In such cases, directly modeling hyperspectral data in the original high-dimensional space while ignoring their manifold characteristics often leads to unstable representations and degraded discriminative structures, resulting in significant performance deterioration and limited generalization capability \cite{li2025back}. Consequently, how to construct stable, robust, and degradation-adaptive models for HSI classification under composite degradation scenarios has become an active and important research topic.

With respect to HSI classification, a large body of research has proposed discriminative deep learning frameworks, including spatial–spectral feature learning methods based on convolutional neural networks (CNNs) \cite{2025Efficient,10843260}, attention mechanisms \cite{rs14091968,9684381}, and Transformers \cite{ahmad2025comprehensive,10976442}. These approaches typically achieve high classification accuracy on standard clean datasets; however, they often rely on an implicit assumption that the training and testing samples share relatively consistent data distributions and structural characteristics. In composite degradation scenarios, this assumption is easily violated, as degradation-induced structural perturbations can cause significant distribution shifts, leading to unstable decision boundaries, reduced intra-class compactness, and aggravated inter-class confusion. Since most discriminative models primarily depend on supervised signals to enhance class separability, the learned representations tend to emphasize \textit{label alignment} rather than \textit{structural alignment}. Consequently, such methods exhibit pronounced vulnerability when confronted with geometric distortions introduced by composite degradations \cite{rs14205199}.

To enhance robustness, existing studies employ perturbation-based strategies like  data augmentation \cite{8877854}, feature reconstruction \cite{9439796}, and noise suppression \cite{10445289}. Representative approaches include HSI-DeNet \cite{8435923} for spatial–spectral denoising, NBAN \cite{9376953} for adaptive band attention, MTGAN \cite{9125995} for joint reconstruction–classification, and HyperTTA \cite{hypertta} for test-time adaptation. However, these methods face two fundamental limitations. First, they typically rely on constrained degradation parameters, failing to cover the continuous composite degradation space encountered in real-world scenarios \cite{hypertta,8542643}. Second, most strategies expand observation-level diversity without enforcing geometric stability in the representation space. Consequently, learned representations often suffer from performance fluctuations when composite degradations drive samples away from their intrinsic low-dimensional manifold \cite{9408225}.

Many current approaches such as self-supervised learning \cite{chen2025spectral,10193882} and contrastive learning \cite{11050958,zhang2025graph} attempt to mitigate distribution shifts via alignment constraints. However, their effectiveness is limited under composite degradations, as they fail to guarantee stable local geometric relationships when multiple factors are superimposed \cite{10459061}. Conversely, diffusion models have emerged for explicit distribution modeling \cite{Spectraldiff,zhang2024data} but mostly operate in the raw observation space \cite{lu2025beyond}. Under composite degradations, this approach is inefficient: it forces the model to capture extensive non-discriminative perturbations alongside the intrinsic structure \cite{qu2024mtlsc}, amplifying structural noise and instability due to the high dimensionality of HSI data \cite{hu2025}. Consequently, explicit manifold constraints are essential to fully realize the potential of diffusion models.

A comprehensive analysis indicates that, under composite degradation conditions, HSI classification has reached a clear bottleneck when relying solely on enhanced discriminative capacity or increased model complexity. The superposition of multiple degradation factors disrupts the assumption that hyperspectral samples are distributed around a unified and compact low-dimensional manifold, thereby substantially increasing the difficulty of modeling spatial–spectral distributions. In such scenarios, the key to robust classification lies not in improving resistance to individual noise types, but in explicitly characterizing and preserving the low-dimensional manifold structure of hyperspectral data under degradation perturbations. By effectively separating high-dimensional redundancy and non-discriminative disturbances, stable classification performance with strong generalization capability can be achieved.

To this end, we propose MSDiff, a unified framework that synergizes discriminative manifold learning with generative diffusion to tackle the challenges of composite degradations. Grounded in the low-dimensional manifold assumption, MSDiff disentangles degradation-induced noise from the intrinsic discriminative structures of HSI by learning a geometrically stable manifold embedding and refining representations via a latent-space diffusion model, thus enabling accurate and consistent classification across diverse degradation conditions. The primary contributions of our approach can be summarized in three aspects:

\begin{itemize}

\item {This paper proposes MSDiff, a robust HSI classification framework tailored for composite degradation scenarios. By explicitly embedding high-dimensional spatial–spectral observations contaminated by complex degradations into a low-dimensional manifold and subsequently employing a generative diffusion model for unified distribution modeling, the proposed framework enables effective classification under severe degradation conditions.}

\item {To address complex high-dimensional perturbations, we construct a discriminative spatial–spectral embedding network guided by joint cross-entropy and reconstruction objectives. This network projects observations corrupted by composite degradations into a compact low-dimensional manifold, effectively reducing representation complexity while preserving essential class-discriminative information. The resulting embedding provides a tractable and well-structured geometric foundation, which is critical for explicitly modeling degradation processes in the latent space.}

\item {To mitigate the complex residual noise and perturbations persisting in the low-dimensional space, we introduce a diffusion model to perform unified modeling of spectral-spatial distributions directly within the manifold. By leveraging generative learning to further refine and smooth the latent representations, this process yields structured and regularized manifold embeddings, thereby achieving accurate HSI classification even under complex composite degradation conditions.}

\end{itemize}

\section{Methodology}

\begin{figure*}[htbp]
	\centering 
	\includegraphics[width=0.9\textwidth]{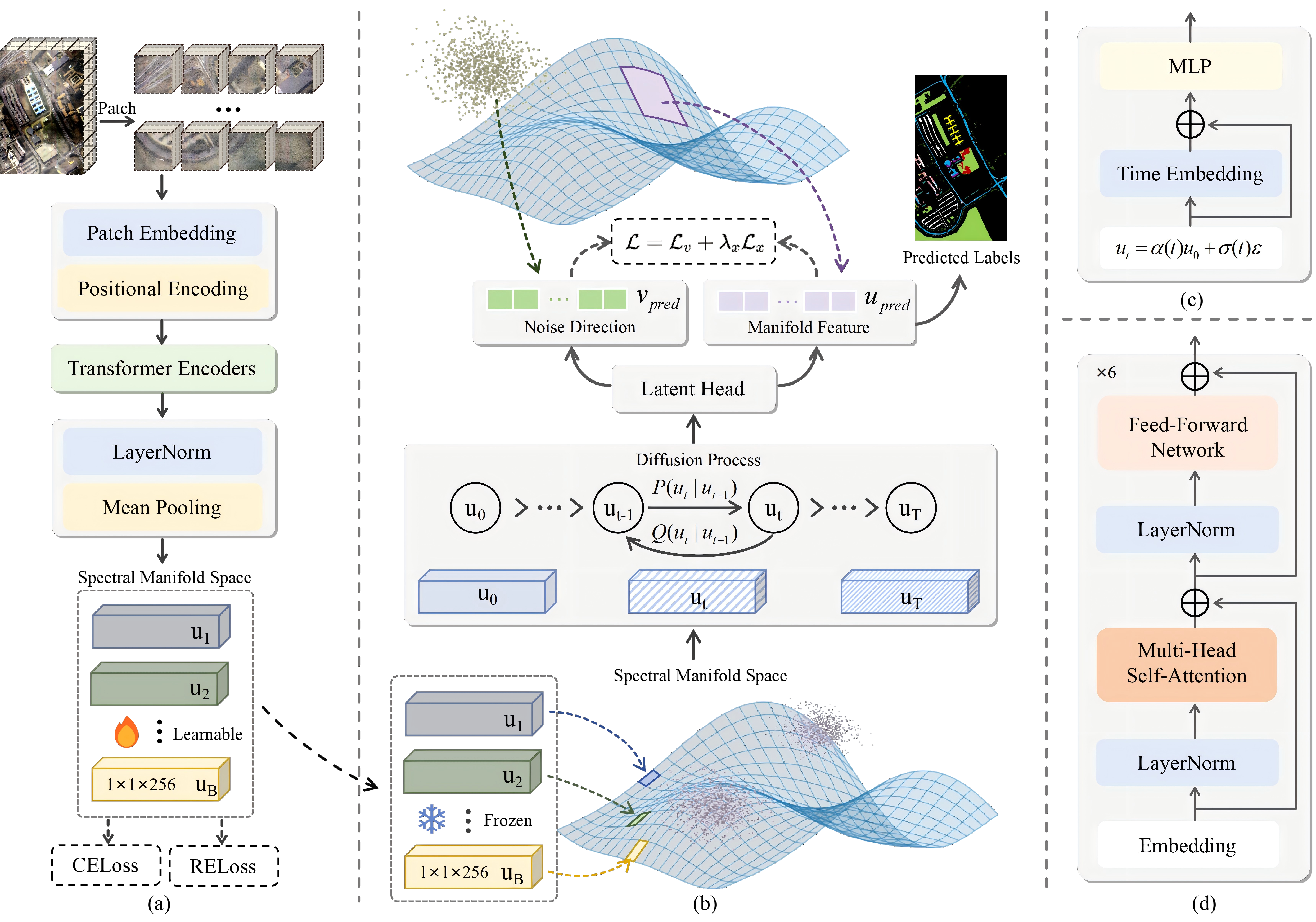} 
	\caption{Overview of the proposed manifold-space diffusion framework. (a) Discriminative low-dimensional spectral–spatial manifold embedding under composite degradations. (b) Diffusion-based generative modeling and refinement in the manifold space for robust HSI classification. $u_0$ and $u_t$ represent manifold coordinate $\mathbf{u}\in\mathbb{R}^{D}$ of time step 0 and time step $T$, respectively. $P(\cdot|\cdot)$ and $Q(\cdot|\cdot)$ represent the forward and reverse diffusion process, respectively. (c) Structure of latent head. (d) Structure of Transformer encoders}
	\label{fig:framework} 
	
\end{figure*}

\subsection{Overview}
As illustrated in Fig.~\ref{fig:framework}, the proposed framework addresses degraded HSI classification by progressively transforming high-dimensional noise-corrupted observations into a compact and discriminative low-dimensional manifold representation. Built upon this manifold embedding, a diffusion-based generative model is further introduced to explicitly regularize and refine the spectral–spatial distributions in the manifold space, enabling robust and consistent classification under complex composite degradation scenarios.

\subsection{Manifold Embedding with Discriminative Constraint}
This model aims to learn a compact latent representation that captures the intrinsic spectral–spatial structure of hyperspectral data while remaining robust to degradation-induced perturbations, as shown in Fig.~\ref{fig:framework} (a). Specifically, a discriminatively constrained manifold embedding network is constructed to project degraded HSI patches into a low-dimensional latent space, where both spectral–spatial reconstruction fidelity and class-discriminative consistency are jointly enforced. The resulting latent representation provides a structured and compact manifold space that facilitates subsequent generative modeling of degradation effects. 

\subsubsection{Composite Degradation Construction}
Let $\mathbf{x}\in\mathbb{R}^{H\times W\times C}$ denote a clean HSI with $C$ spectral bands. Under practical remote sensing conditions, the observed image is often contaminated by multiple degradation sources acting jointly, including sensor noise, atmospheric interference, blur, impulse corruption, and compression artifacts. We model a composite degradation process as in Eq.~\eqref{1}:
\begin{eqnarray}
	\tilde{\mathbf{x}}=\mathcal{D}(\mathbf{x};w,\rho)
	\label{1}
\end{eqnarray}
where $\mathcal{D}(\cdot)$ denotes a degradation operator, $w\in\mathbb{R}^{K}$ is a Dirichlet-sampled mixing weight over $K$ degradation types, and $\rho$ controls the overall degradation intensity.

This formulation yields a continuously varying composite degradation space, rather than a fixed set of predefined corruption types. Given a degraded HSI patch $\tilde{\mathbf{x}}\in\mathbb{R}^{P\times P\times C}$, the objective of this work is to learn a discriminative representation that remains stable under composite degradations, while preserving class-separable structure for classification.

\subsubsection{Low-rank Inductive and Manifold Hypothesis}
HSI data exhibit strong spectral correlation and spatial redundancy, and are widely regarded as high-dimensional observations lying on a low-dimensional latent manifold. Formally, we assume that there exists an unknown nonlinear mapping in Eq.~\eqref{2}
\begin{eqnarray}
	\Phi:\mathbb{R}^{P\times P\times C}\to\mathcal{M}\subset\mathbb{R}^D,\quad D\ll P^2C
	\label{2}
\end{eqnarray}
such that clean samples from the same class concentrate on a compact region of a latent manifold $\mathcal{M}$.

However, under composite degradations, the observed samples $\tilde{\mathbf{x}}$ are displaced away from $\mathcal{M}$, yielding distorted representations in high-dimensional feature space and significantly degrading classification performance. Therefore, explicitly recovering a compact, discriminative manifold representation from degraded observations is a critical prerequisite for robust HSI classification.

\subsubsection{Discriminative Manifold Embedding}
To operationalize the above hypothesis, we design a bottlenecked spectral–spatial embedding network that maps degraded HSI patches into a compact latent space. Given an input patch $\tilde{\mathbf{x}}$, we first perform spectral–spatial patchification and linear projection, as shown in Eq.~\eqref{3}:
\begin{eqnarray}
	\mathbf{z}_0=\text{PatchEmbed}(\tilde{\mathbf{x}})\in\mathbb{R}^{N\times D}
	\label{3}
\end{eqnarray}
where $N=(P/s)^{2}$ is the number of patches with stride $s$, and $D$ is the embedding dimension. The patch embedding module contains an explicit low-rank bottleneck along the spectral dimension, enforcing dimensionality reduction before entering the transformer blocks.

The embedded tokens are then processed by a stack of transformer blocks, as shown in Eq.~\eqref{4}:
\begin{eqnarray}
	\mathbf{z}_\ell=\mathcal{T}_\ell(\mathbf{z}_{\ell-1}),\quad\ell=1,\ldots,L
	\label{4}
\end{eqnarray}
where each block $\mathcal{T}_{\ell}(\cdot)$ consists of multi-head self-attention and feed-forward layers with RMS normalization.

Finally, the latent manifold coordinate $\mathbf{u}\in\mathbb{R}^{D}$ is obtained by token aggregation, as shown in Eq.~\eqref{5}:
\begin{eqnarray}
	\mathbf{u}=\frac{1}{N}\sum_{i=1}^{N}\mathrm{Norm}(\mathbf{z}_{L}^{(i)})
	\label{5}
\end{eqnarray}
which serves as a compact representation of the spectral–spatial manifold coordinate corresponding to the degraded input patch.

Although the bottleneck operation is applied at the embedding stage, the manifold coordinate $\mathbf{u}$ is extracted after deep contextual interaction, ensuring that it encodes global spectral–spatial semantics rather than shallow statistics.

\subsubsection{Discriminative and Reconstruction-guided Manifold Constraint}
A compact latent manifold alone is insufficient to guarantee robustness under composite degradations, as manifolds learned solely through discriminative supervision may favor label separability while lacking geometric stability against degradation-induced perturbations. To address this issue, we impose a joint reconstruction–discrimination constraint on the manifold embedding, explicitly regularizing its geometric structure. 

Given a degraded input patch $\tilde{\mathbf{x}}$, the backbone network produces a latent manifold coordinate $\mathbf{u}\in\mathbb{R}^{D}$. A lightweight decoder is then applied to reconstruct the spectral–spatial signal, as shown in Eq.~\eqref{6}:
\begin{eqnarray}
	\hat{\mathbf{x}}=\mathcal{G}(\mathbf{u})
	\label{6}
\end{eqnarray}
where $\mathcal{G}(\cdot)$ denotes a learnable reconstruction head.

The reconstruction loss is defined as in Eq.~\eqref{7}:
\begin{eqnarray}
	\mathcal{L}_{\mathrm{rec}}=\left\|\hat{\mathbf{x}}-\mathbf{x}\right\|_2^2
	\label{7}
\end{eqnarray}
where $\mathbf{x}$ denotes the corresponding clean reference patch.

This objective encourages the latent coordinate $\mathbf{u}$ to retain essential spectral–spatial information while suppressing degradation-specific noise components, thereby pulling degraded samples back toward the underlying data manifold. To further enforce semantic consistency, a classification head is applied directly on the manifold coordinate, as shown in Eq.~\eqref{8}:
\begin{eqnarray}
	\hat{\mathbf{y}}=\mathbf{W}_c\mathbf{u}+\mathbf{b}_c
	\label{8}
\end{eqnarray}
and optimized using the cross-entropy loss, as shown in Eq.~\eqref{9}:
\begin{eqnarray}
	\mathcal{L}_{\mathrm{cls}}=-\sum_{c=1}^Cy_c\log\mathrm{Softmax}(\hat{\mathbf{y}})_c
	\label{9}
\end{eqnarray}
Unlike purely reconstruction-based objectives, this discriminative constraint explicitly aligns the learned manifold structure with class boundaries, ensuring that samples from different land-cover categories remain separable in the latent space.

The final training objective of the manifold embedding network is given in Eq.~\eqref{10}: 
\begin{eqnarray}
\mathcal{L}_{\mathrm{embed}}=\mathcal{L}_{\mathrm{rec}}+\lambda_{\mathrm{cls}}\mathcal{L}_{\mathrm{cls}},
	\label{10}
\end{eqnarray}
where $\lambda_{\mathrm{cls}}$ controls the trade-off between reconstruction fidelity and discriminative power.

Through joint optimization, the latent representation $\mathbf{u}$ naturally acquires a structured manifold geometry: reconstruction supervision encourages smoothness and continuity, while classification supervision imposes semantic partitioning. We refer to the resulting representation space as a spectral–spatial manifold space, which serves as the foundation for subsequent diffusion-based distribution modeling.
\subsection{Diffusion-based Generative Modeling}
Although the embedding in Section II-B creates a compact discriminative manifold, it remains a task-oriented approximation where degraded samples deviate from the ideal geometry due to structured perturbations. To regularize these deviations, we introduce a generative diffusion process directly within the learned manifold space, as illustrated in Fig.~\ref{fig:framework} (b). Unlike high-dimensional spectral diffusion, operating on compact latent coordinates explicitly models degradation dynamics while preserving the semantic structure captured by the discriminative embedding.  

\subsubsection{Manifold-Space Diffusion}
Specifically, the forward diffusion process, denoted as $P(\mathbf{u}_t\mid\mathbf{u}_{t-1})$, gradually injects Gaussian noise into the manifold representation through a predefined variance schedule $\{\beta_t\}_{t=1}^T$, as shown in Eq.~\eqref{111}:
\begin{eqnarray}
P(\mathbf{u}_t\mid\mathbf{u}_{t-1})=\mathcal{N}\left(\mathbf{u}_t;\sqrt{1-\beta_t}\mathbf{u}_{t-1},\beta_t\mathbf{I}\right)
	\label{111}
\end{eqnarray}
which results in a closed-form expression, as shown in Eq.~\eqref{11}:
\begin{eqnarray}
\mathbf{u}_t=\alpha(t)\mathbf{u}_0+\sigma(t)\boldsymbol{\varepsilon},\quad\boldsymbol{\varepsilon}\sim\mathcal{N}(\mathbf{0},\mathbf{I})
	\label{11}
\end{eqnarray}
where $t\in[0,1]$ is a continuous diffusion time variable. The noise schedule is parameterized by Eq.~\eqref{12}:
\begin{eqnarray}
\alpha(t)=\cos\left(\frac{\pi t}{2}\right),\quad\sigma(t)=\sin\left(\frac{\pi t}{2}\right)
	\label{12}
\end{eqnarray}
which smoothly interpolates between the clean manifold representation at $t=0$ and pure Gaussian noise at $t=1$.

The reverse diffusion process, denoted as $Q(\mathbf{u}_{t-1}\mid\mathbf{u}_t)$, aims to progressively remove noise and recover the clean manifold structure. Since the true reverse distribution is intractable, we parameterize it using a neural network conditioned on the diffusion timestep $t$, as shown in Eq.~\eqref{122}:
\begin{eqnarray}
Q_\theta(\mathbf{u}_{t-1}\mid\mathbf{u}_t)=\mathcal{N}\left(\mathbf{u}_{t-1};\boldsymbol{\mu}_\theta(\mathbf{u}_t,t),\boldsymbol{\Sigma}_\theta(t)\right)
	\label{122}
\end{eqnarray}

Compared with conventional pixel-space diffusion, this formulation explicitly assumes that degradation primarily induces perturbations orthogonal to the intrinsic data manifold, while the semantic structure is encoded in the low-dimensional coordinate $\mathbf{u}_0$. Consequently, diffusion in the manifold space provides a principled mechanism to disentangle degradation-induced variability from intrinsic class-discriminative information.

\subsubsection{Manifold-Aware Generative Modeling}
To reverse the diffusion process and recover the clean manifold representation, we design a lightweight latent diffusion head conditioned on the diffusion time $t$. Given a noisy manifold coordinate $\mathbf{u}_t$, the diffusion head jointly predicts both the noise component and the clean manifold coordinate, as shown in Eq.~\eqref{13}:
\begin{eqnarray}
(\mathbf{v}_{\mathrm{pred}},\mathbf{u}_{\mathrm{pred}})=f_\theta(\mathbf{u}_t,t)
	\label{13}
\end{eqnarray}
where $\mathbf{v}_{\mathrm{pred}}$ denotes the predicted noise direction, and $\mathbf{u}_{\mathrm{pred}}$ represents the estimated clean manifold coordinate. The diffusion head $f_{\theta}(\cdot)$ is implemented as a multi-layer perceptron augmented with a sinusoidal time embedding, which encodes the diffusion timestep as shown in Eq.~\eqref{14}:
\begin{eqnarray}
\gamma(t)=\left[\sin(\omega_1t),\cos(\omega_1t),\ldots,\sin(\omega_Kt),\cos(\omega_Kt)\right]
	\label{14}
\end{eqnarray}
with logarithmically spaced frequencies $\{\omega_k\}_{k=1}^K$, enabling the model to adaptively modulate its behavior across different noise levels.

Unlike standard noise-prediction-only diffusion formulations, we adopt a manifold-aware x-prediction strategy that explicitly reconstructs the clean manifold representation $\mathbf{u}_{\mathrm{pred}}$. This design aligns with the geometric interpretation of hyperspectral manifolds: the diffusion process is encouraged not only to estimate degradation-induced stochasticity, but also to project degraded manifold samples back onto a compact and intrinsic low-dimensional manifold. 

The latent diffusion model is optimized using a composite objective function, as shown in Eq.~\eqref{15} and~\eqref{155} :
\begin{eqnarray}
\mathcal{L}_{\text{diff}} = \mathcal{L}_v + \lambda_x \mathcal{L}_x
	\label{15}
\end{eqnarray}
\begin{eqnarray}
\mathcal{L}_v = \mathbb{E}_{t,\boldsymbol{\epsilon}} \big[ \lVert \mathbf{v}_{\text{pred}} - \boldsymbol{\epsilon} \rVert_2^2 \big], \quad
\mathcal{L}_x = \mathbb{E}_{t} \big[ \lVert \mathbf{u}_{\text{pred}} - \mathbf{u}_0 \rVert_2^2 \big]
	\label{155}
\end{eqnarray}
where $\mathcal{L}_{v}$ enforces accurate prediction of the injected noise, and $\mathcal{L}_{x}$ explicitly constrains the reconstructed manifold feature to align with the clean latent coordinate $\mathbf{u}_0$. In practice, $\lambda_{x}$ is set to 1, yielding a balanced formulation that jointly accounts for stochastic modeling and manifold fidelity.

From a geometric perspective, the noise prediction term enables the diffusion head to model local stochastic variations around the learned manifold, while the manifold reconstruction term enforces global structural consistency by pulling perturbed samples back toward the task-oriented low-dimensional manifold. Rather than acting as a conventional denoiser in latent space, the proposed diffusion process serves as a geometry-aware regularizer that smooths the manifold structure and suppresses degradation-induced distortions, resulting in stable and robust representations for HSI classification under composite degradation scenarios.

\subsubsection{Regularized Manifold Learning for Classification}
After training, the diffusion model serves as a manifold refinement module. Given a latent representation $\mathbf{u}_{\mathrm{raw}}$ extracted from a composite-degraded sample, we obtain a diffusion-regularized manifold feature by evaluating the diffusion head at a fixed intermediate time $t=t^{*}$, as shown in Eq.~\eqref{16}:
\begin{eqnarray}
\mathbf{u}_{\mathrm{diff}}=\hat{\mathbf{u}}_0=f_\theta\left(\alpha(t^*)\mathbf{u}_{\mathrm{raw}},t^*\right)
	\label{16}
\end{eqnarray}
This operation can be interpreted as a single-step projection from a degraded manifold neighborhood back toward the clean manifold.

The refined representation $\mathbf{u}_{\mathrm{diff}}$ is subsequently fed into a lightweight classifier for HSI classification. Notably, the backbone encoder and diffusion model are frozen during this stage, ensuring that the classifier operates on a stable, diffusion-regularized manifold space. This design decouples generative modeling from discriminative training and avoids overfitting to specific degradation patterns.

\section{Experiments}
\subsection{Experimental Settings}
\subsubsection{Datasets}

In order to verify the effectiveness of our algorithm, we applied the algorithm to two public datasets: the Pavia University (PU) dataset and the WHU-Hi-LongKou (WHLK) dataset.

Collected by the ROSIS sensor over Pavia, Italy, the PU dataset has a spatial resolution of 1.3 $m$, dimensions of $610\times340$ pixels, and 103 usable spectral bands (430–860 $nm$) after removing 12 noisy channels from the original 115; it annotates nine land-cover categories, serves as a standard HSI classification benchmark due to its spatial–spectral complexity.


Acquired via a UAV-borne Headwall Nano-Hyperspec sensor over Longkou, China, the WHLK dataset \cite{8573977,ZHONG2020112012} has a $550\times400$ pixel size, 0.463 $m$ spatial resolution and 270 spectral bands spanning 400–1000 nm; depicting a heterogeneous agricultural scene with nine land-cover classes, it is widely recognized as a rigorous benchmark for HSI classification in complex scenarios.


\subsubsection{Single Degradation Scenarios}
To simulate various HSI degradation phenomena commonly encountered in real-world scenarios, nine types of degraded data were generated for this study using the Multi-Degradation Simulator (MDS) proposed in the HyperTTA \cite{hypertta} framework.

\begin{itemize}
\item   JPEG Compression: Caused by lossy storage or transmission, this degradation leads to detail loss and blocky artifacts, especially in high-frequency regions. 
\item Zero-Mean Gaussian Noise: Caused by sensor thermal effects or electromagnetic interference, this noise affects all spectral bands and is simulated using an adjustable standard deviation $\sigma$ to control intensity.
\item Additive Gaussian Noise: Independent of the original signal, this is simulated by adding zero-mean noise with adjustable variance to each normalized spectral band, where $\sigma$ controls the degradation intensity.
\item Poisson Noise: Poisson noise originates from the quantum randomness of photon detection and is commonly observed in HSI under low-light or nighttime conditions.
\item Salt \& Pepper Noise: A typical impulse degradation caused by sensor faults or transmission errors, characterized by randomly distributed extreme bright and dark pixels.
\item Stripe Noise: A common degradation caused by sensor failures, transmission errors, or scanning artifacts, manifesting as regularly spaced vertical stripes. 
\item Deadline Noise: A stripe-like degradation caused by sensor malfunction or transmission errors, characterized by the complete loss of pixel columns in certain spectral bands.
\item Convolutional Blur: This type of degradation typically arises from low-pass filtering during the sensor imaging process, lens defocus, or information loss during data transmission.
\item Fog Degradation: This degradation primarily results from the scattering and absorption of light by water vapor and suspended particles in the atmosphere. 

\end{itemize}

\subsubsection{Composite Degradation Scenarios}
\begin{figure}[htbp]
	\centering 
	\includegraphics[width=0.4\textwidth]{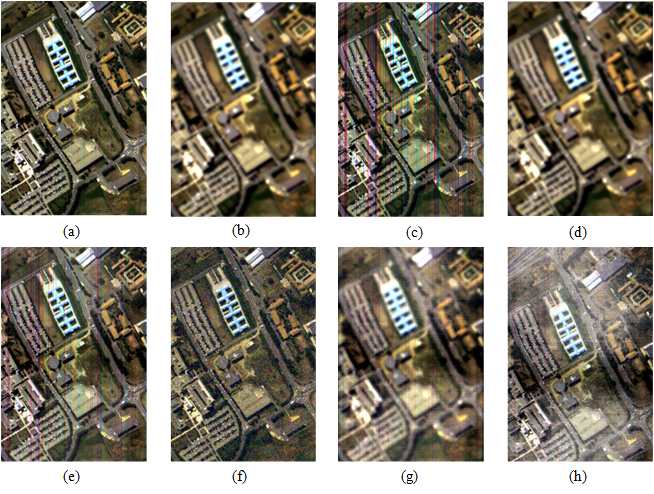} 
	\caption{Visualization of HSI with composite degradations on PU datasets.} 
	\label{fig:composite} 
\end{figure}

\begin{table}[htbp]
\small
\centering
\caption{Definition and notation of composite degradation scenarios used for evaluation.}
\label{tab:comlabel}
\setlength{\tabcolsep}{2mm}
\renewcommand{\arraystretch}{1.1}
\resizebox{\columnwidth}{!}{%
\begin{tabular}{c|c|r l|r}
\specialrule{1.2pt}{0pt}{0pt}
\multirow{2}{*}{Fig.~\ref{fig:composite}} &
\multirow{2}{*}{\textbf{Complexity}} &
\multicolumn{2}{l|}{\multirow{2}{*}{\textbf{Original Description}}} &
\multirow{2}{*}{\textbf{Label}}\\
 & & \multicolumn{2}{c|}{} & \\
\midrule\midrule
(a) & 3-mix & \multicolumn{2}{l|}{Deadlines + Poisson + Salt and pepper} & C-3-1\\
(b) & 3-mix & \multicolumn{2}{l|}{Jpeg + Blur + Fog} & C-3-2\\
(c) & 3-mix & \multicolumn{2}{l|}{Additive + Stripes + Zmgauss} & C-3-3\\
(d) & 3-mix & \multicolumn{2}{l|}{Poisson + kernel + Fog} & C-3-4\\
(e) & 5-mix & \multicolumn{2}{l|}{Deadlines + Stripes + Blur + Salt and pepper + Fog} & C-5-1\\
(f) & 5-mix & \multicolumn{2}{l|}{Jpeg + Additive + Poisson + Zmgauss + Salt and pepper} & C-5-2\\
(g) & 7-mix & \multicolumn{2}{l|}{all except Additive and Stripes} & C-7\\
(h) & 9-mix & \multicolumn{2}{l|}{all degradations} & C-9\\
\specialrule{1.2pt}{0pt}{0pt}
\end{tabular}}
\end{table}

To better approximate realistic scenarios, we construct a Composite Degradation Space (CDS), in which multiple degradation types are simultaneously imposed on clean hyperspectral observations. Rather than treating different degradations independently, the proposed setting explicitly models their joint effects, enabling a more comprehensive assessment of classification robustness under compounded perturbations.

Fig.~\ref{fig:composite} illustrates representative examples of composite degradations generated on the PU dataset. Specifically, we consider four levels of composite complexity, corresponding to three-mix, five-mix, seven-mix, and nine-mix degradation configurations. 

For clarity and conciseness in subsequent quantitative analysis, we adopt a unified notation to denote different composite degradation scenarios, as shown in Table~\ref{tab:comlabel}. Each test case is denoted as C-k-i, where \textit{k} indicates the number of degradation types involved in the composite configuration, and \textit{i} denotes the index of the corresponding scenario at the same complexity level. For example, C-3-1 represents the first three-mix composite degradation case shown in Fig.~\ref{fig:composite}, while C-9 corresponds to the full nine-degradation composite scenario. This notation is consistently used throughout the remainder of the paper. 

During training, composite degradations are dynamically generated within a continuous CDS using randomly sampled Dirichlet weights, encouraging the model to learn representations invariant to varying degradation mixtures. For fair and reproducible evaluation, the testing phase employs a fixed set of standardized benchmarks as shown in Table~\ref{tab:comlabel}, which range from three-mix to nine-mix configurations and provide a controlled subset of the degradation space for systematic comparison. 


\subsubsection{Evaluation Metrics}
We will compare the effectiveness of our algorithm with other algorithms from three aspects, mainly including Overall Accuracy (OA), Average Accuracy (AA), kappa coefficient ($\kappa$).

\subsubsection{Implementation Details}
All experiments were conducted on a workstation equipped with an Intel® Xeon® Platinum 8336C CPU @ 2.30GHz, 64 cores, 128GB of memory, and two NVIDIA GeForce RTX 4090 GPUs, each with 24GB of memory. The proposed model was implemented using the Pytorch framework. 

For all experiments, HSI were processed patch-wise with a $25\times25$ patch size for datasets to balance local spatial context and computational efficiency; the datasets were randomly split by pixel coordinates into 10\% training, 10\% validation and 80\% test sets, with the same split applied across all experiments to ensure fair comparisons. The backbone network, based on the JiT hyperspectral transformer \cite{li2025back}, was first trained under composite degradation conditions to learn compact latent manifold representations, optimized via AdamW with a $1\times10^{-3}$ learning rate and $1\times10^{-4}$ weight decay; after convergence, backbone parameters were frozen, and a lightweight diffusion head was trained on the learned latent manifold coordinates to model degradation-induced perturbations, adopting a cosine noise schedule with continuous time sampling and a combined v-parameterization and x-prediction objective, and optimized using AdamW with identical learning rate and weight decay settings.

For classification, the diffusion-regularized latent representations were extracted by applying the trained diffusion model at a fixed diffusion time, and a shallow multilayer perceptron classifier was trained on top of these refined features. During this stage, both the backbone and diffusion head were kept frozen, and only the classifier parameters were updated. Cross-entropy loss was used for supervision.

\subsection{Performance Analysis}

\begin{table*}[htbp]
\centering
\small
\caption{Comparison of classification performance on the PU dataset under different degradation types.}
\label{tab:pu_results}
\begin{adjustbox}{width=1.0\textwidth}
\begin{tabular}{
  c|c| 
  C{1.05cm}C{1.4cm}C{1.05cm}C{1.4cm}C{1.05cm}C{1.05cm}
  C{1.05cm}C{1.05cm}C{1.05cm}C{1.05cm}C{1.05cm}C{1.05cm}C{1.05cm}C{1.05cm} 
  C{1.05cm} 
}
\toprule
\textbf{Method} & \textbf{Metric} & \textbf{Jpeg} & \textbf{\footnotesize Zero-Mean} & \textbf{Additive}  & \textbf{\footnotesize Salt\&Pepper} & \textbf{Stripes} & \textbf{Deadline} & \textbf{C-3-1} & \textbf{C-3-2} & \textbf{C-3-3} & \textbf{C-3-4} & \textbf{C-5-1} & \textbf{C-5-2} & \textbf{C-7} & \textbf{C-9} & \textbf{Avg.} \\
\midrule \midrule
\multirow{3}{*}{HyperTTA}
& OA (\%)           & 88.32 & 75.71 & 89.15 & 85.32 & 94.43 & 73.24 & 97.51 & 55.83 & 97.61 & 55.82 & 52.97 & 88.51 & 51.95 & 60.10 & 76.18 \\
& AA (\%)           & 81.13 & 60.67 & 83.83 & 76.42 & 89.98 & 71.11 & 97.22 & 41.97 & 97.43 & 42.25 & 31.96 & 83.54 & 33.02 & 41.02 & 66.54 \\
& Kappa$\times$100  & 84.07 & 67.07 & 85.15 & 80.06 & 92.63 & 65.71 & 96.70 & 31.16 & 96.84 & 31.50 & 31.48 & 84.35 & 22.28 & 45.09 & 65.29 \\
\midrule
\multirow{3}{*}{SSEFN}
& OA (\%)           & 69.25 & 13.40 & 20.09 & 15.40 & 21.32 & 16.65 & 64.29 & 71.16 & 25.42 & 71.18 & 35.90 & 22.17 & 59.43 & 46.69 & 39.45 \\
& AA (\%)           & 76.43 & 12.94 & 20.26 & 15.16 & 23.13 & 17.96 & 75.73 & 83.17 & 27.12 & 83.10 & 38.77 & 22.88 & 63.86 & 48.88 & 43.53 \\
& Kappa$\times$100  & 62.54 & 1.71 & 10.08 & 3.79 & 11.05 & 6.16 & 57.38 & 65.18 & 15.36 & 65.2 & 26.55 & 12.20 & 51.43 & 37.83 & 30.46 \\
\midrule
\multirow{3}{*}{SpectralDiff}
& OA (\%)           & 71.64 & 56.38 & 63.09 & 62.74 & 89.51 & 89.18 & 98.76 & 93.04 & 98.91 & 93.77 & 74.62 & 65.33 & 82.37 & 77.67 & 79.79 \\
& AA (\%)           & 61.19 & 40.93 & 46.52 & 42.32 & 86.04 & 85.58 & 98.08 & 90.87 & 98.29 & 92.08 & 72.91 & 49.43 & 77.51 & 67.13 & 72.06 \\
& Kappa$\times$100  & 58.28 & 38.22 & 44.86 & 44.19 & 85.80 & 85.34 & 98.36 & 90.67 & 98.55 & 91.66 & 72.89 & 48.34 & 75.81 & 69.69 & 71.62 \\
\midrule
\multirow{3}{*}{SLCGC}
& OA (\%)           & 64.79 & 47.51 & 48.25 & 48.67 & 42.70 & 21.83 & 63.86 & 50.25 & 45.58 & 50.44 & 44.84 & 50.21 & 41.14 & 57.87 & 48.42 \\
& AA (\%)           & 60.15 & 50.53 & 52.54 & 52.82 & 50.45 & 29.05 & 59.97 & 46.96 & 51.24 & 45.12 & 52.28 & 54.31 & 41.14 & 58.08 & 50.33 \\
& Kappa$\times$100  & 54.64 & 37.24 & 38.51 & 38.96 & 32.27 & 10.53 & 54.13 & 38.75 & 35.50 & 39.31 & 35.45 & 40.63 & 30.10 & 49.85 & 38.28 \\
\midrule
\multirow{3}{*}{RPCL-FSL}
& OA (\%)           & 60.44 & 12.39 & 15.26 & 15.12 & 35.01 & 23.78 & 58.73 & 48.86 & 37.25 & 48.62 & 36.84 & 29.12 & 35.09 & 21.67 & 34.16 \\
& AA (\%)           & 61.28 & 15.10 & 27.44 & 24.08 & 46.91 & 35.44 & 63.65 & 52.29 & 54.54 & 52.06 & 53.16 & 47.53 & 43.48 & 34.29 & 43.66 \\
& Kappa$\times$100  & 49.95 & 3.28 & 7.07 & 6.14 & 24.58 & 14.65 & 49.19 & 35.90 & 30.08 & 35.56 & 27.84 & 20.76 & 22.63 & 12.56 & 24.30 \\
\midrule
\multirow{3}{*}{SS-MTr}
& OA (\%)           & 50.31 & 46.74 & 45.85 & 47.41 & 56.9 & 38.11 & 84.21 & 58.53 & 64.06 & 58.50 & 74.80 & 61.60 & 56.22 & 52.18 & 56.82 \\
& AA (\%)           & 24.30 & 22.18 & 20.43 & 23.74 & 62.32 & 15.14 & 94.82 & 48.80 & 73.83 & 48.74 & 85.28 & 49.13 & 41.57 & 49.17 & 47.10 \\
& Kappa$\times$100  & 30.47 & 7.99 & 22.29 & 9.55 & 44.75 & 9.69 & 80.32 & 41.9 & 55.26 & 42.01 & 68.92 & 41.74 & 38.38 & 42.08 & 38.24 \\
\midrule
\multirow{3}{*}{DSFormer}
& OA (\%)           & 45.75 & 10.26 & 12.80 & 14.63 & 19.18 & 17.90 & 37.02 & 45.77 & 24.58 & 37.62 & 3.77 & 4.09 & 4.72 & 8.34 & 20.46 \\
& AA (\%)           & 54.86 & 28.17 & 28.68 & 35.92 & 38.55 & 32.79 & 17.02 & 42.35 & 12.76 & 29.42 & 8.15 & 6.31 & 9.67 & 15.45 & 25.72 \\
& Kappa$\times$100  & 36.71 & 6.60 & 7.78 & 9.95 & 13.31 & 11.02 & 20.82 & 29.96 & 13.34 & 20.11 & -5.69 & -5.90 & -4.90 & -0.45 & 10.90 \\
\midrule
\multirow{3}{*}{\textbf{Ours}}
& OA (\%)           & \textbf{99.52} & \textbf{94.11} & \textbf{99.52} & \textbf{98.33} & \textbf{99.10} & \textbf{98.65} & \textbf{99.50} & \textbf{97.79} & \textbf{99.40} & \textbf{95.54} & \textbf{83.12} & \textbf{99.40} & \textbf{83.65} & \textbf{93.06} & \textbf{95.76} \\
& AA (\%)           & \textbf{99.35} & \textbf{93.12} & \textbf{99.34} & \textbf{97.92} & \textbf{99.02} & \textbf{98.75} & \textbf{99.33} & \textbf{96.85} & \textbf{99.20} & \textbf{93.92} & \textbf{79.33} & \textbf{99.19} & \textbf{80.83} & \textbf{91.83} & \textbf{94.86} \\
& Kappa$\times$100  & \textbf{99.37} & \textbf{92.24} & \textbf{99.36} & \textbf{97.79} & \textbf{98.81} & \textbf{98.21} & \textbf{99.34} & \textbf{97.05} & \textbf{99.21} & \textbf{93.99} & \textbf{78.94} & \textbf{99.20} & \textbf{78.39} & \textbf{90.83} & \textbf{94.48} \\
\bottomrule
\end{tabular}
\end{adjustbox}
\end{table*}

\begin{table*}[htbp]
\centering
\small
\caption{Comparison of classification performance on the WHLK dataset under different degradation types.}
\label{tab:lk_results}
\begin{adjustbox}{width=1.0\textwidth}
\begin{tabular}{
  c|c| 
  C{1.05cm}C{1.4cm}C{1.05cm}C{1.4cm}C{1.05cm}C{1.05cm}
  C{1.05cm}C{1.05cm}C{1.05cm}C{1.05cm}C{1.05cm}C{1.05cm}C{1.05cm}C{1.05cm} 
  C{1.05cm} 
}
\toprule
\textbf{Method} & \textbf{Metric} & \textbf{Jpeg} & \textbf{\footnotesize Zero-Mean} & \textbf{Additive}  & \textbf{\footnotesize Salt\&Pepper} & \textbf{Stripes} & \textbf{Deadline} & \textbf{C-3-1} & \textbf{C-3-2} & \textbf{C-3-3} & \textbf{C-3-4} & \textbf{C-5-1} & \textbf{C-5-2} & \textbf{C-7} & \textbf{C-9} & \textbf{Avg.} \\
\midrule \midrule
\multirow{3}{*}{HyperTTA}
& OA (\%)           & 87.72 & 97.08 & 98.70 & 97.25 & 99.10 & 65.16 & 51.88 & 69.56 & 99.14 & 64.58 & 55.12 & 67.02 & 5.90 & 7.88 & 69.01 \\
& AA (\%)           & 93.29 & 90.01 & 95.81 & 90.38 & 98.36 & 76.66 & 68.84 & 71.60 & 98.33 & 69.77 & 69.83 & 77.60 & 14.55 & 24.69 & 74.27 \\
& Kappa$\times$100  & 85.32 & 96.15 & 98.29 & 96.38 & 98.82 & 58.98 & 44.61 & 63.21 & 98.87 & 57.94 & 48.73 & 60.35 & 3.07 & 4.89 & 65.40 \\
\midrule
\multirow{3}{*}{SSEFN}
& OA (\%)           & 63.04 & 20.46 & 14.58 & 21.93 & 57.71 & 21.93 & 65.18 & 94.07 & 66.14 & 93.63 & 76.99 & 30.10 & 79.78 & 72.76 & 55.59 \\
& AA (\%)           & 57.33 & 14.51 & 18.73 & 16.08 & 41.80 & 17.09 & 49.06 & 93.54 & 49.28 & 93.08 & 61.05 & 26.93 & 66.82 & 57.87 & 47.37 \\
& Kappa$\times$100  & 54.38 & 5.88 & 1.60 & 7.21 & 47.33 & 10.39 & 55.82 & 92.32 & 57.12 & 91.75 & 70.45 & 16.70 & 74.08 & 65.41 & 46.46 \\
\midrule
\multirow{3}{*}{SpectralDiff}
& OA (\%)           & 93.96 & 71.13 & 74.43 & 63.11 & 97.78 & 97.56 & 90.37 & 91.53 & 97.04 & 44.98 & 94.32 & 80.04 & 95.36 & 67.75 & 82.81 \\
& AA (\%)           & 84.30 & 37.74 & 40.89 & 29.50 & 92.65 & 92.14 & 77.72 & 93.34 & 90.12 & 36.63 & 94.61 & 47.45 & 84.99 & 39.98 & 67.29 \\
& Kappa$\times$100  & 91.98 & 59.62 & 64.70 & 49.08 & 97.07 & 96.79 & 87.30 & 88.76 & 96.09 & 21.24 & 97.79 & 72.75 & 93.86 & 54.52 & 76.54 \\
\midrule
\multirow{3}{*}{SLCGC}
& OA (\%)           & 72.03 & 73.33 & 70.78 & 74.86 & 70.71 & 72.74 & 71.34 & 67.62 & 70.38 & 67.93 & 69.87 & 72.04 & 51.87 & 53.76 & 68.52 \\
& AA (\%)           & 46.91 & 45.26 & 44.28 & 47.24 & 53.12 & 50.65 & 47.30 & 51.64 & 48.52 & 52.15 & 54.61 & 46.14 & 43.79 & 45.27 & 48.35 \\
& Kappa$\times$100  & 64.52 & 65.45 & 62.52 & 67.36 & 63.25 & 65.45 & 63.51 & 59.94 & 63.79 & 60.30 & 62.35 & 64.07 & 42.95 & 44.88 & 60.74 \\
\midrule
\multirow{3}{*}{RPCL-FSL}
& OA (\%)           & 77.95 & 24.49 & 58.09 & 37.07 & 73.44 & 58.43 & 71.48 & 80.14 & 72.54 & 80.18 & 74.03 & 40.61 & 68.38 & 57.52 & 62.45 \\
& AA (\%)           & 69.46 & 20.45 & 46.09 & 28.28 & 65.62 & 54.74 & 61.20 & 67.41 & 65.60 & 67.65 & 63.85 & 36.78 & 54.52 & 42.19 & 53.13 \\
& Kappa$\times$100  & 72.18 & 16.54 & 49.33 & 27.41 & 66.93 & 50.11 & 64.41 & 74.73 & 65.81 & 74.78 & 67.17 & 31.54 & 60.12 & 47.24 & 54.88 \\
\midrule
\multirow{3}{*}{SS-MTr}
& OA (\%)           & 75.74 & 39.07 & 59.09 & 52.63 & 66.62 & 63.72 & 34.09 & 94.66 & 82.55 & 93.25 & 82.80 & 84.29 & 29.88 & 23.25 & 62.97 \\
& AA (\%)           & 77.05 & 29.09 & 53.39 & 37.75 & 74.92 & 66.96 & 15.28 & 93.64 & 86.84 & 87.69 & 78.54 & 81.44 & 28.50 & 19.30 & 59.31 \\
& Kappa$\times$100  & 70.06 & 29.62 & 50.72 & 42.58 & 60.06 & 55.92 & 2.98 & 93.04 & 78.16 & 91.18 & 77.78 & 79.97 & 19.72 & 11.15 & 54.50 \\
\midrule
\multirow{3}{*}{DSFormer}
& OA (\%)           & 57.99 & 37.71 & 39.88 & 35.38 & 40.21 & 39.36 & 6.95 & 7.69 & 5.24 & 7.70 & 5.53 & 5.46 & 5.57 & 5.92 & 21.47 \\
& AA (\%)           & 44.33 & 24.53 & 24.31 & 24.90 & 28.24 & 29.23 & 20.59 & 21.75 & 18.81 & 21.76 & 20.06 & 21.12 & 18.87 & 15.53 & 23.86 \\
& Kappa$\times$100  & 47.80 & 27.21 & 27.84 & 25.28 & 29.27 & 28.69 & 4.07 & 5.05 & 2.44 & 5.03 & 2.85 & 2.98 & 3.11 & 3.09 & 15.34 \\
\midrule
\multirow{3}{*}{\textbf{Ours}}
& OA (\%)           & \textbf{99.37} & \textbf{98.70} & \textbf{99.19} & \textbf{99.05} & \textbf{99.47} & \textbf{98.19} & \textbf{99.59} & \textbf{98.86} & \textbf{99.50} & \textbf{99.23} & \textbf{96.33} & \textbf{95.77} & \textbf{98.88} & \textbf{97.56} & \textbf{98.55} \\
& AA (\%)           & \textbf{98.78} & \textbf{96.99} & \textbf{97.03} & \textbf{98.16} & \textbf{98.47} & \textbf{95.87} & \textbf{99.23} & \textbf{97.61} & \textbf{98.63} & \textbf{98.14} & \textbf{94.52} & \textbf{96.03} & \textbf{97.48} & \textbf{96.15} & \textbf{97.36} \\
& Kappa$\times$100  & \textbf{99.17} & \textbf{98.29} & \textbf{98.94} & \textbf{98.75} & \textbf{99.30} & \textbf{97.63} & \textbf{99.46} & \textbf{98.50} & \textbf{99.34} & \textbf{98.98} & \textbf{95.19} & \textbf{94.40} & \textbf{95.53} & \textbf{96.80} & \textbf{97.88} \\

\bottomrule
\end{tabular}
\end{adjustbox}
\end{table*}

To evaluate the effectiveness of the proposed method, a variety of representative algorithms were selected as comparative experiments. Specifically, SpectralDiff \cite{Spectraldiff} serves as a diffusion-based baseline, while SS-MTr \cite{ssmtr} and DSFormer \cite{DSFormer} represent recent Transformer-based models for spectral–spatial feature modeling. SLCGC \cite{slcgc}, RPCL-FSL \cite{RPCL}, and SSEFN \cite{SSEFN} are included as self-supervised or weakly supervised baselines that exploit intrinsic spectral–spatial structures under limited supervision. In addition, HyperTTA\cite{hypertta} represents a recent model specifically designed to handle degraded data and has shown promising performance in such scenarios. Note that, all models were trained under identical conditions on clean data and evaluated across various simulated degradation scenarios on the PU and WHLK datasets. 

\subsubsection{Quantitative Results}
\begin{figure}[!h]
	\centering 
	\includegraphics[width=0.30\textwidth]{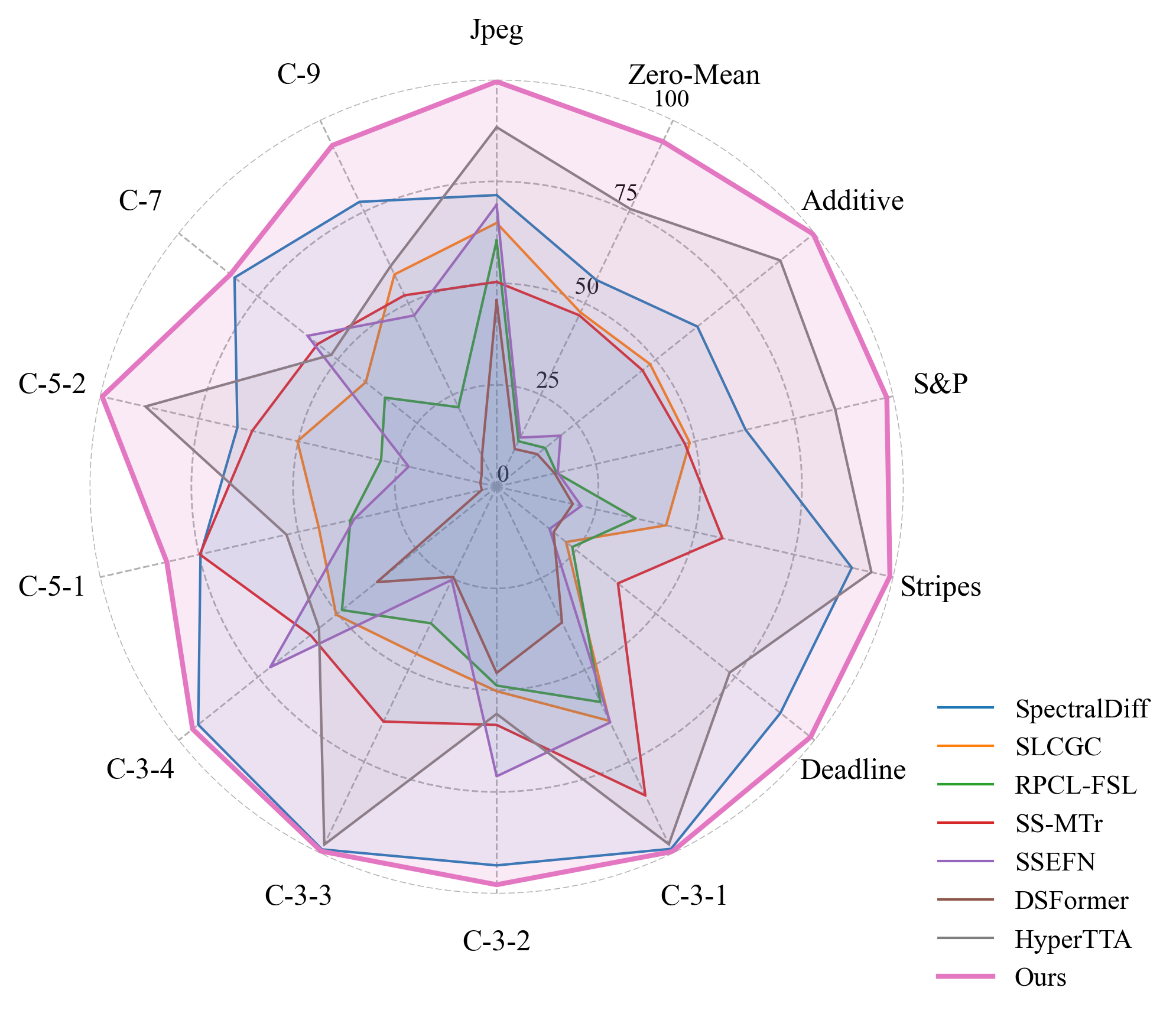} 
	\caption{Radar chart of classification performance (OA) on the PU dataset under different types of degradation.} 
	\label{fig:radarpu} 
\end{figure}

\begin{figure}[!h]
	\centering 
	\includegraphics[width=0.30\textwidth]{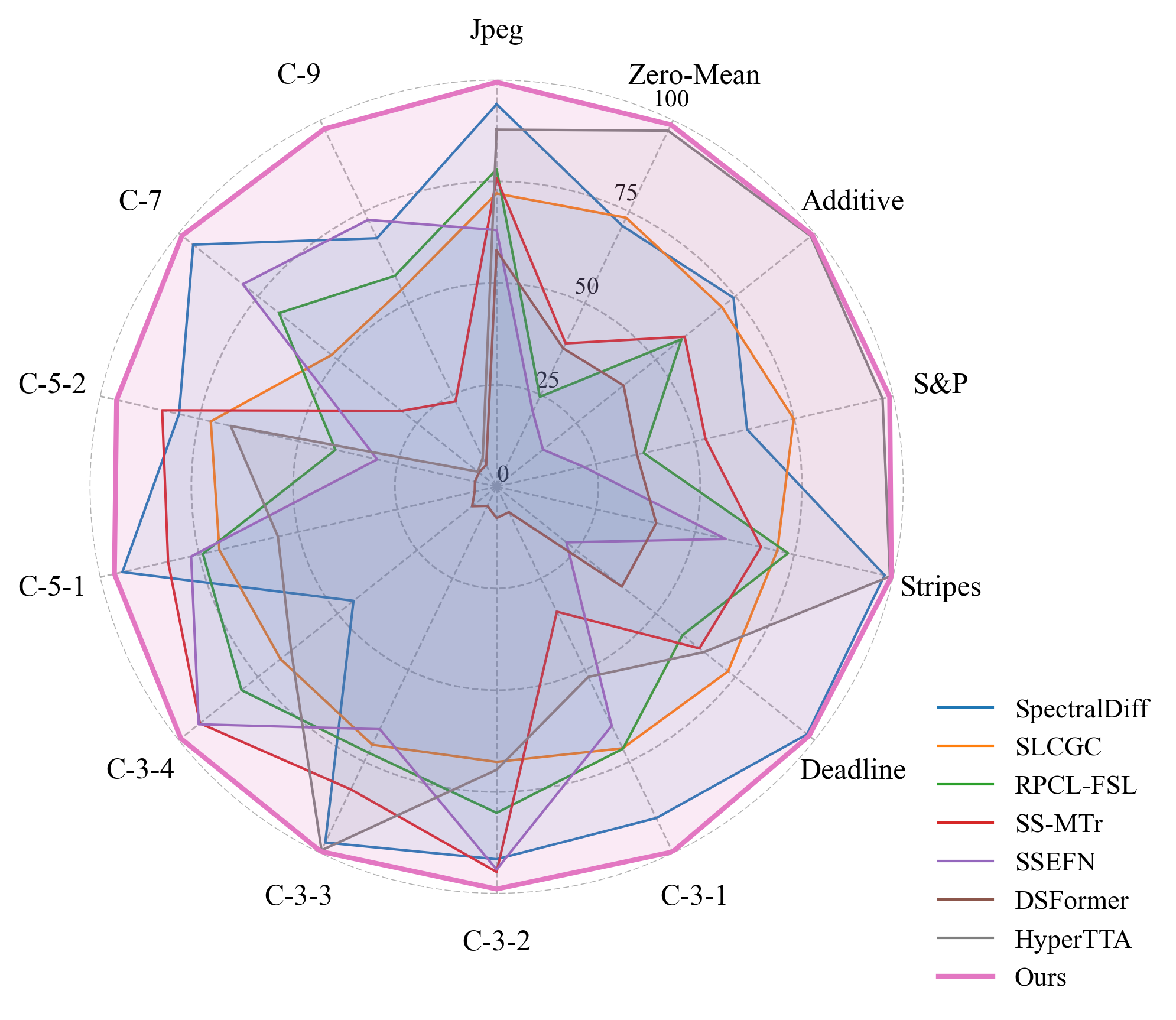} 
	\caption{Radar chart of classification performance (OA) on the WHLK dataset under different types of degradation.} 
	\label{fig:radarlk} 
\end{figure}

Tables~\ref{tab:pu_results} and \ref{tab:lk_results} report the quantitative classification results on the PU and WHLK datasets, respectively, under six representative single-degradation scenarios and eight composite degradation scenarios defined in Section III-A. The visualized OA performance is illustrated in Figs.~\ref{fig:radarpu} and \ref{fig:radarlk}.

Under single-degradation conditions, baselines like SpectralDiff and HyperTTA exhibit degradation-dependent robustness, performing adequately on structured artifacts (e.g., stripes) but degrading significantly under random noise (e.g., Gaussian). In contrast, MSDiff maintains consistently high accuracy across all settings on both datasets. This robustness stems from the proposed low-dimensional manifold embedding, which constrains representations to a compact latent space, effectively suppressing high-dimensional perturbations while preserving discriminative spectral–spatial structures. 

As shown in tables~\ref{tab:pu_results} and \ref{tab:lk_results}, baseline performance deteriorates significantly as composite complexity escalates from three-mix to nine-mix settings. While sporadic improvements suggest implicit noise cancellation, such instability indicates a failure to learn truly invariant representations. Conversely, MSDiff demonstrates remarkable stability across all levels, maintaining OA values above 93\% on PU and 97\% on WHLK even under the challenging C-9 scenario. This confirms the efficacy of enforcing representation compactness within the low-dimensional manifold to mitigate complex perturbations.

The superior performance of MSDiff under both single and composite degradations can be explained from a manifold perspective. Composite degradations introduce complex, coupled perturbations that displace hyperspectral observations away from their intrinsic low-dimensional manifold in the high-dimensional spectral–spatial space. Traditional discriminative models tend to overfit to specific degradation patterns and fail to maintain stable decision boundaries under such compounded distortions. By contrast, MSDiff explicitly embeds degraded observations into a compact low-dimensional manifold and further refines this representation through diffusion-based generative modeling. This design allows the model to progressively smooth degradation-induced perturbations while preserving class-discriminative structures on the manifold. As a result, MSDiff achieves robust and consistent classification performance, even when multiple degradation sources jointly act on the data.

These quantitative results clearly demonstrate that manifold-aware generative modeling provides a principled and effective solution for HSI classification under complex composite degradation scenarios, significantly outperforming existing degradation-robust and diffusion-based baselines. 

\subsubsection{Visual Evaluation}

\begin{figure*}
	\centering
	\includegraphics[width=0.75\linewidth]{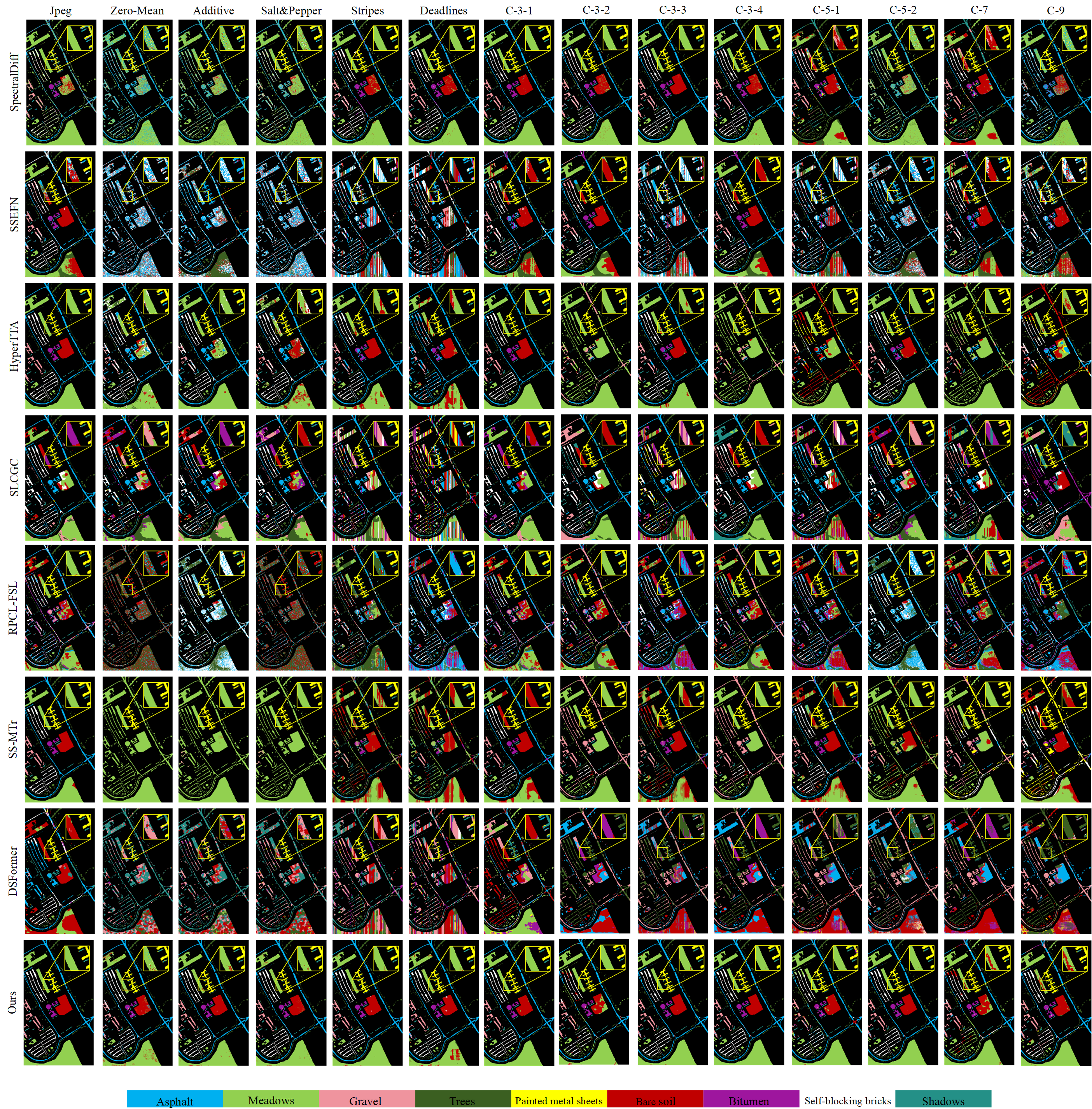}
	\caption{The visual classification results of the PU dataset under different degradation types using various model algorithms. Each row corresponds to a specific method, and each column represents a particular type of degradation.}
	\label{fig:PU}
\end{figure*}

\begin{figure*}
	\centering
	\includegraphics[width=1.0\linewidth]{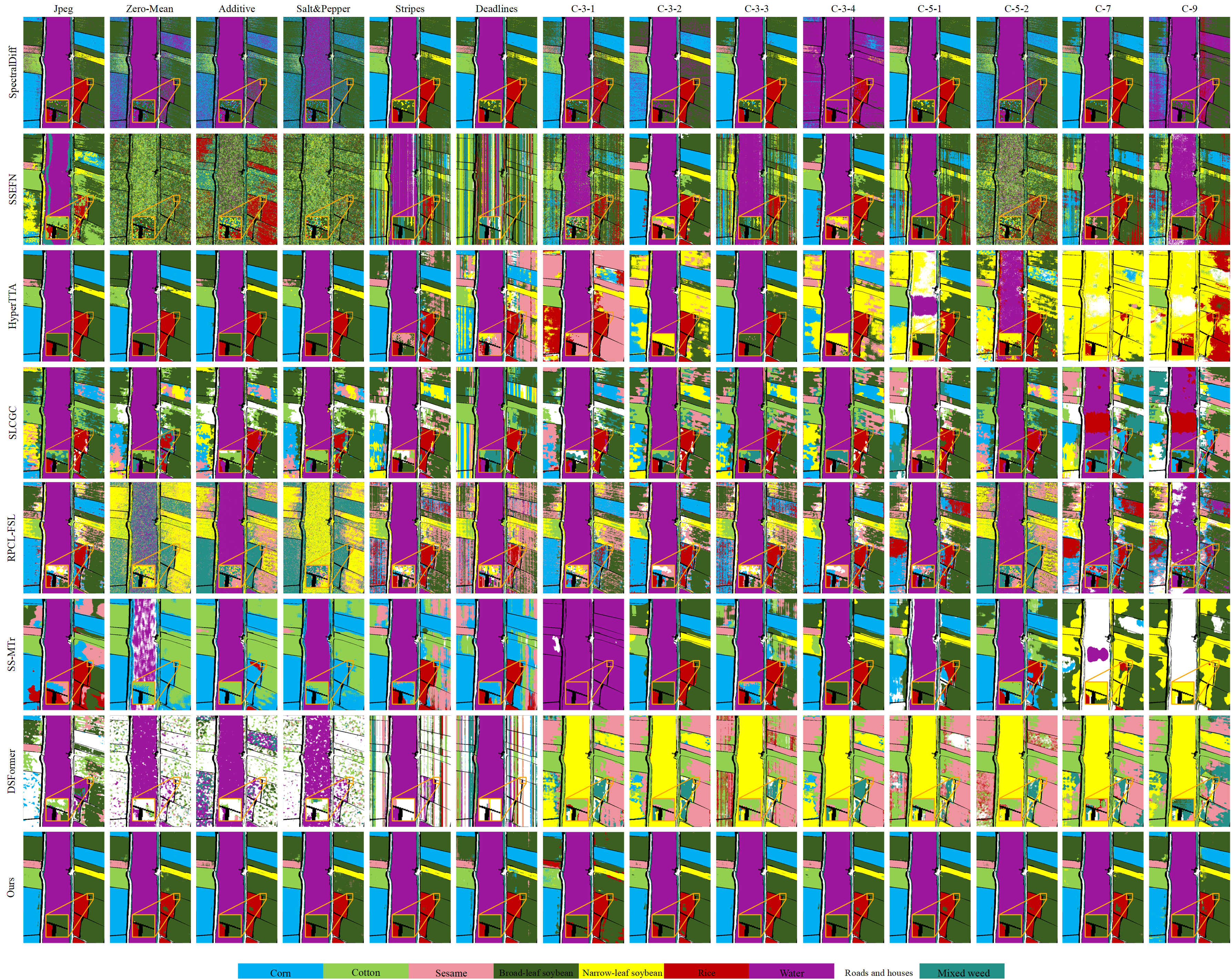}
	\caption{The visual classification results of the WHLK dataset under different degradation types using various model algorithms. Each row corresponds to a specific method, and each column represents a particular type of degradation.}
	\label{fig:WHLK}
\end{figure*}

Fig.~\ref{fig:PU} and Fig.~\ref{fig:WHLK} present the visual classification results on the PU and WHLK datasets, respectively, under both single and composite degradation scenarios. To facilitate a clearer inspection of local structural details and classification boundaries, representative regions of interest are selected and displayed in a magnified manner for both datasets. As observed, most baseline methods suffer from pronounced spatial inconsistency and class fragmentation, and these issues become more severe under composite degradation conditions. As the degradation complexity increases from three-mix to seven-mix and nine-mix configurations, many methods produce a large amount of noise-like misclassifications within homogeneous regions and exhibit evident discontinuities along class boundaries, indicating a high sensitivity to compounded degradation perturbations.

Although certain methods exhibit apparent visual stability under composite degradations, this is often a byproduct of over-smoothing and reduced discriminative capability rather than genuine robustness. Such superficial smoothness comes at the cost of essential structural details, leading to poor region integrity and compromised boundary accuracy, particularly in fine-grained areas. 

In contrast, MSDiff consistently produces spatially coherent, boundary-preserving, and semantically consistent classification results across all single and composite degradation scenarios. Even under the most challenging nine-mix composite degradation, MSDiff effectively suppresses noise interference, maintains the integrity of homogeneous regions, and accurately delineates boundaries between different land-cover classes. This stable visual performance is highly consistent with the quantitative results reported earlier, demonstrating that the proposed method can effectively absorb high-dimensional perturbations induced by composite degradations within a low-dimensional manifold space, and further smooth the spectral–spatial distributions through diffusion-based modeling, thereby enabling robust HSI classification under complex degradation conditions.

\subsection{Model Analysis}
\subsubsection{Ablation Study}

\begin{table*}[htbp]
\centering
\small
\caption{Ablation study on Manifold and Diffusion modules on PU dataset.}
\label{tab:pu_ablation_results}
\begin{adjustbox}{width=1.0\textwidth}
\begin{tabular}{
  c|c| 
  C{1.05cm}C{1.4cm}C{1.05cm}C{1.4cm}C{1.05cm}C{1.05cm}
  C{1.05cm}C{1.05cm}C{1.05cm}C{1.05cm}C{1.05cm}C{1.05cm}C{1.05cm}C{1.05cm} 
  C{1.05cm} 
}
\toprule
\textbf{Method} & \textbf{Metric} & \textbf{Jpeg} & \textbf{\footnotesize Zero-Mean} & \textbf{Additive}  & \textbf{\footnotesize Salt\&Pepper} & \textbf{Stripes} & \textbf{Deadline} & \textbf{C-3-1} & \textbf{C-3-2} & \textbf{C-3-3} & \textbf{C-3-4} & \textbf{C-5-1} & \textbf{C-5-2} & \textbf{C-7} & \textbf{C-9} & \textbf{Avg.} \\
\midrule \midrule
\multirow{3}{*}{\textbf{Ours}}
& OA (\%)           & \textbf{99.52} & \textbf{94.11} & \textbf{99.52} & \textbf{98.33} & \textbf{99.10} & \textbf{98.65} & \textbf{99.50} & \textbf{97.79} & \textbf{99.40} & \textbf{95.54} & \textbf{83.12} & \textbf{99.40} & \textbf{83.65} & \textbf{93.06} & \textbf{95.76} \\
& AA (\%)           & \textbf{99.35} & \textbf{93.12} & \textbf{99.34} & \textbf{97.92} & \textbf{99.02} & \textbf{98.75} & \textbf{99.33} & \textbf{96.85} & \textbf{99.20} & \textbf{93.92} & \textbf{79.33} & \textbf{99.19} & \textbf{80.83} & \textbf{91.83} & \textbf{94.86} \\
& Kappa$\times$100  & \textbf{99.37} & \textbf{92.24} & \textbf{99.36} & \textbf{97.79} & \textbf{98.81} & \textbf{98.21} & \textbf{99.34} & \textbf{97.05} & \textbf{99.21} & \textbf{93.99} & \textbf{78.94} & \textbf{99.20} & \textbf{78.39} & \textbf{90.83} & \textbf{94.48} \\
\midrule
\multirow{3}{*}{w/o Manifold}
& OA (\%)           & 42.46 & 53.51 & 64.00 & 78.10 & 62.70 & 71.55 & 84.75 & 81.55 & 85.60 & 20.12 & 69.65 & 69.60 & 59.48 & 63.88 & 64.78 \\
& AA (\%)           & 17.70 & 49.14 & 69.76 & 72.84 & 47.80 & 47.34 & 83.22 & 79.76 & 84.62 & 77.93 & 69.49 & 69.52 & 54.85 & 61.34 & 63.24 \\
& Kappa$\times$100  & 24.38 & 50.12 & 66.14 & 69.04 & 54.58 & 63.03 & 82.94 & 78.50 & 85.47 & 76.49 & 69.54 & 69.47 & 55.51 & 61.87 & 64.79 \\
\midrule
\multirow{3}{*}{w/o Diffusion}
& OA (\%)           & 99.20 & 83.97 & 95.55 & 92.43 & 98.84 & 89.41 & 92.66 & 92.37 & 95.25 & 91.34 & 72.75 & 78.92 & 82.18 & 90.70 & 89.68 \\
& AA (\%)           & 98.87 & 84.22 & 96.31 & 90.41 & 87.39 & 90.08 & 94.94 & 90.82 & 89.32 & 87.91 & 67.02 & 78.25 & 79.86 & 90.94 & 87.60 \\
& Kappa$\times$100  & 98.55 & 75.06 & 94.19 & 87.13 & 96.47 & 86.80 & 90.41 & 91.97 & 87.77 & 89.67 & 65.27 & 70.81 & 77.56 & 88.73 & 85.74 \\
\bottomrule
\end{tabular}
\end{adjustbox}
\end{table*}

\begin{table*}[htbp]
\centering
\small
\caption{Ablation study on Manifold and Diffusion modules on WHLK dataset.}
\label{tab:lk_ablation_results}
\begin{adjustbox}{width=1.0\textwidth}
\begin{tabular}{
  c|c| 
  C{1.05cm}C{1.4cm}C{1.05cm}C{1.4cm}C{1.05cm}C{1.05cm}
  C{1.05cm}C{1.05cm}C{1.05cm}C{1.05cm}C{1.05cm}C{1.05cm}C{1.05cm}C{1.05cm} 
  C{1.05cm} 
}
\toprule
\textbf{Method} & \textbf{Metric} & \textbf{Jpeg} & \textbf{\footnotesize Zero-Mean} & \textbf{Additive}  & \textbf{\footnotesize Salt\&Pepper} & \textbf{Stripes} & \textbf{Deadline} & \textbf{C-3-1} & \textbf{C-3-2} & \textbf{C-3-3} & \textbf{C-3-4} & \textbf{C-5-1} & \textbf{C-5-2} & \textbf{C-7} & \textbf{C-9} & \textbf{Avg.} \\
\midrule \midrule
\multirow{3}{*}{\textbf{Ours}}
& OA (\%)           & \textbf{99.37} & \textbf{98.70} & \textbf{99.19} & \textbf{99.05} & \textbf{99.47} & \textbf{98.19} & \textbf{99.59} & \textbf{98.86} & \textbf{99.50} & \textbf{99.23} & \textbf{96.33} & \textbf{95.77} & \textbf{98.88} & \textbf{97.56} & \textbf{98.55} \\
& AA (\%)           & \textbf{98.78} & \textbf{96.99} & \textbf{97.03} & \textbf{98.16} & \textbf{98.47} & \textbf{95.87} & \textbf{99.23} & \textbf{97.61} & \textbf{98.63} & \textbf{98.14} & \textbf{94.52} & \textbf{96.03} & \textbf{97.48} & \textbf{96.15} & \textbf{97.36} \\
& Kappa$\times$100  & \textbf{99.17} & \textbf{98.29} & \textbf{98.94} & \textbf{98.75} & \textbf{99.30} & \textbf{97.63} & \textbf{99.46} & \textbf{98.50} & \textbf{99.34} & \textbf{98.98} & \textbf{95.19} & \textbf{94.40} & \textbf{95.53} & \textbf{96.80} & \textbf{97.88} \\
\midrule
\multirow{3}{*}{w/o Manifold}
& OA (\%)           & 99.22 & 56.92 & 94.18 & 56.75 & 99.12 & 86.76 & 99.25 & 85.19 & 73.48 & 78.53 & 56.95 & 71.05 & 72.04 & 58.26 & 77.69 \\
& AA (\%)           & 98.55 & 73.59 & 90.48 & 59.64 & 98.06 & 87.65 & 98.57 & 78.23 & 68.83 & 74.2 & 60.65 & 75.59 & 75.54 & 62.01 & 78.69 \\
& Kappa$\times$100  & 98.98 & 51.56 & 92.35 & 49.06 & 98.85 & 83.08 & 99.02 & 85.43 & 67.17 & 73.46 & 50.32 & 68.18 & 64.25 & 51.73 & 73.82 \\
\midrule
\multirow{3}{*}{w/o Diffusion}
& OA (\%)           & 98.65 & 98.45 & 98.53 & 98.37 & 99.16 & 97.08 & 98.65 & 95.62 & 98.69 & 85.18 & 93.95 & 89.91 & 98.49 & 95.73 & 96.18 \\
& AA (\%)           & 98.10 & 96.86 & 97.40 & 97.44 & 97.92 & 95.62 & 98.13 & 91.50 & 97.54 & 76.78 & 90.78 & 87.12 & 96.59 & 93.61 & 93.96 \\
& Kappa$\times$100  & 98.54 & 97.97 & 98.38 & 97.86 & 99.02 & 96.18 & 98.54 & 91.03 & 98.44 & 76.67 & 92.09 & 88.43 & 98.02 & 94.41 & 94.68 \\

\bottomrule
\end{tabular}
\end{adjustbox}
\end{table*}

To validate the effectiveness of the proposed components in MSDiff, we conduct comprehensive ablation studies on both the PU and WHLK datasets. Specifically, two ablated variants are considered: w/o Manifold, which removes the discriminative low-dimensional manifold embedding and directly performs diffusion-based classification in the original high-dimensional spectral–spatial space; w/o Diffusion, which preserves the low-dimensional manifold embedding but replaces the diffusion-based generative modeling with a simple MLP classifier operating on the manifold representations. The quantitative results are reported in Tables  \ref{tab:pu_ablation_results} and  \ref{tab:lk_ablation_results}

As detailed in Table \ref{tab:pu_ablation_results}, removing the manifold embedding leads to a precipitous performance drop (Avg OA decreases from 95.76\% to 64.78\%), particularly under complex scenarios like C-7 and C-9 where accuracy falls to 60\%. This confirms the instability of modeling composite degradations directly in the high-dimensional space. While the w/o Diffusion variant significantly recovers performance (Avg OA 89.68\%), it still lags behind the full MSDiff, notably in the C-9 case (90.70\% vs. 93.06\%). This gap indicates that while embedding suppresses major perturbations, diffusion-based regularization is essential to address residual distribution irregularities that discriminative classifiers alone cannot handle.

The results on the WHLK dataset in  Table \ref{tab:lk_ablation_results} further validate the proposed components. The exclusion of the manifold embedding results in a drastic average OA decline to 77.69\%, with the C-9 scenario suffering a sharp drop to 58.26\% which exposes the instability of modeling in the high-dimensional space. Furthermore, although the w/o Diffusion variant remains competitive in simpler settings, it exhibits consistent performance degradation under complex C-7 and C-9 conditions compared to MSDiff. This persistent gap underscores the critical role of diffusion modeling in overcoming the limitations of purely discriminative learning within the manifold space.

The ablation studies on both datasets clearly validate the complementary roles of the two proposed components. The low-dimensional manifold embedding with discriminative constraints effectively reduces the complexity of spectral–spatial representations and provides a stable foundation for handling composite degradations. Building upon this foundation, the diffusion-based generative modeling explicitly refines the manifold distribution, smoothing degradation-induced perturbations that cannot be fully addressed by discriminative objectives alone. Their joint integration is therefore essential for achieving robust and consistent HSI classification under complex real-world degradation conditions.

\begin{figure*}
	\centering
	\includegraphics[width=0.75\linewidth]{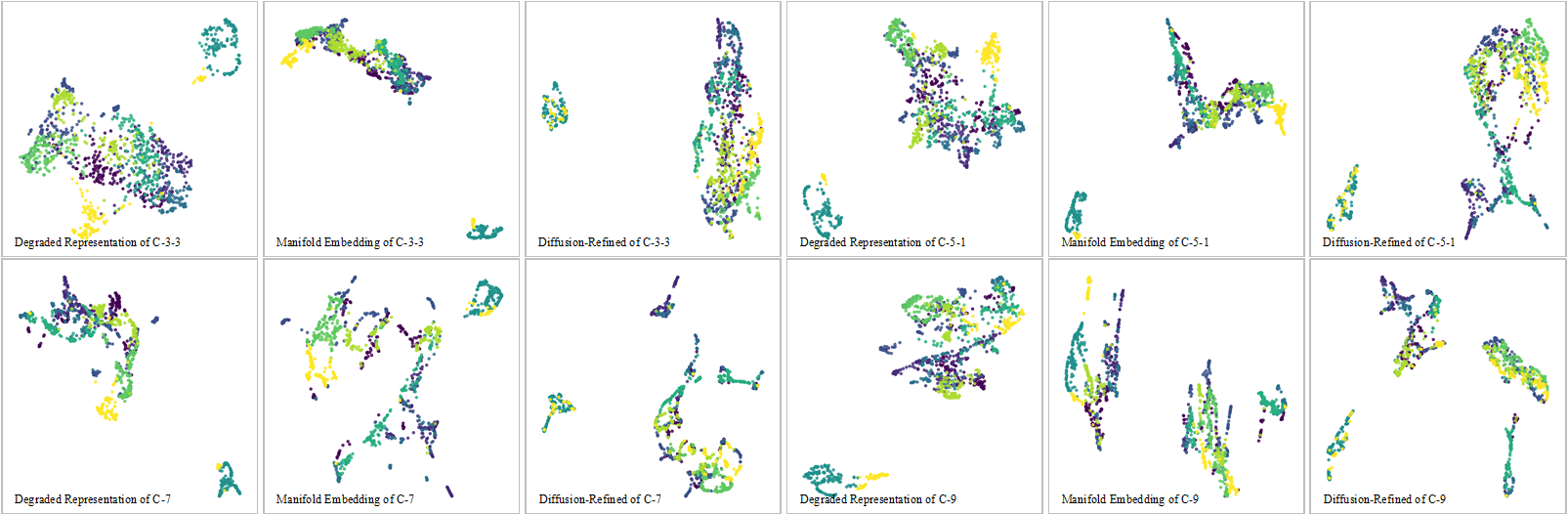}
	\caption{UMAP visualization of representations under different composite degradation levels on PU dataset. }
	\label{fig:UMAP}
\end{figure*}

To qualitatively examine the impact of composite degradations on the intrinsic geometry of hyperspectral data, we visualize the learned representations using UMAP under four composite degradation settings (C-3-3, C-5-1, C-7, and C-9), as shown in Fig.~\ref{fig:UMAP}. In each group, the degraded representation, manifold embedding, and diffusion-refined representation are presented from left to right. Each point corresponds to a sampled hyperspectral pixel, and colors indicate ground-truth class labels for reference.

As observed in Fig.~\ref{fig:UMAP}, composite degradations substantially distort the intrinsic low-dimensional manifold structure of hyperspectral data, scattering and fragmenting samples into irregular clusters in the degraded representation space, with this deviation becoming more pronounced as the number of mixed degradations increases and leading to highly dispersed, geometrically inconsistent embeddings; in contrast, the proposed manifold embedding module enables the representations to exhibit a noticeably more coherent and continuous structure, pulling the data distribution progressively back toward a lower-dimensional manifold despite partial inter-class interleaving, and this effect is further strengthened by diffusion-based refinement, which helps embeddings form smoother, more structured trajectories with reduced geometric fragmentation even under severe composite degradations (e.g., C-9). It is important to note that this visualization focuses on the geometric regularity of the learned representations rather than demonstrating classification separability, as hyperspectral classes often show continuous spectral transitions and overlapping distributions in the intrinsic manifold, and the improved classification performance is achieved by the subsequent classifier exploiting these geometrically regularized features instead of relying on explicit class-wise separation in the embedding space.

\begin{figure}[!h]
	\centering 
	\includegraphics[width=0.35\textwidth]{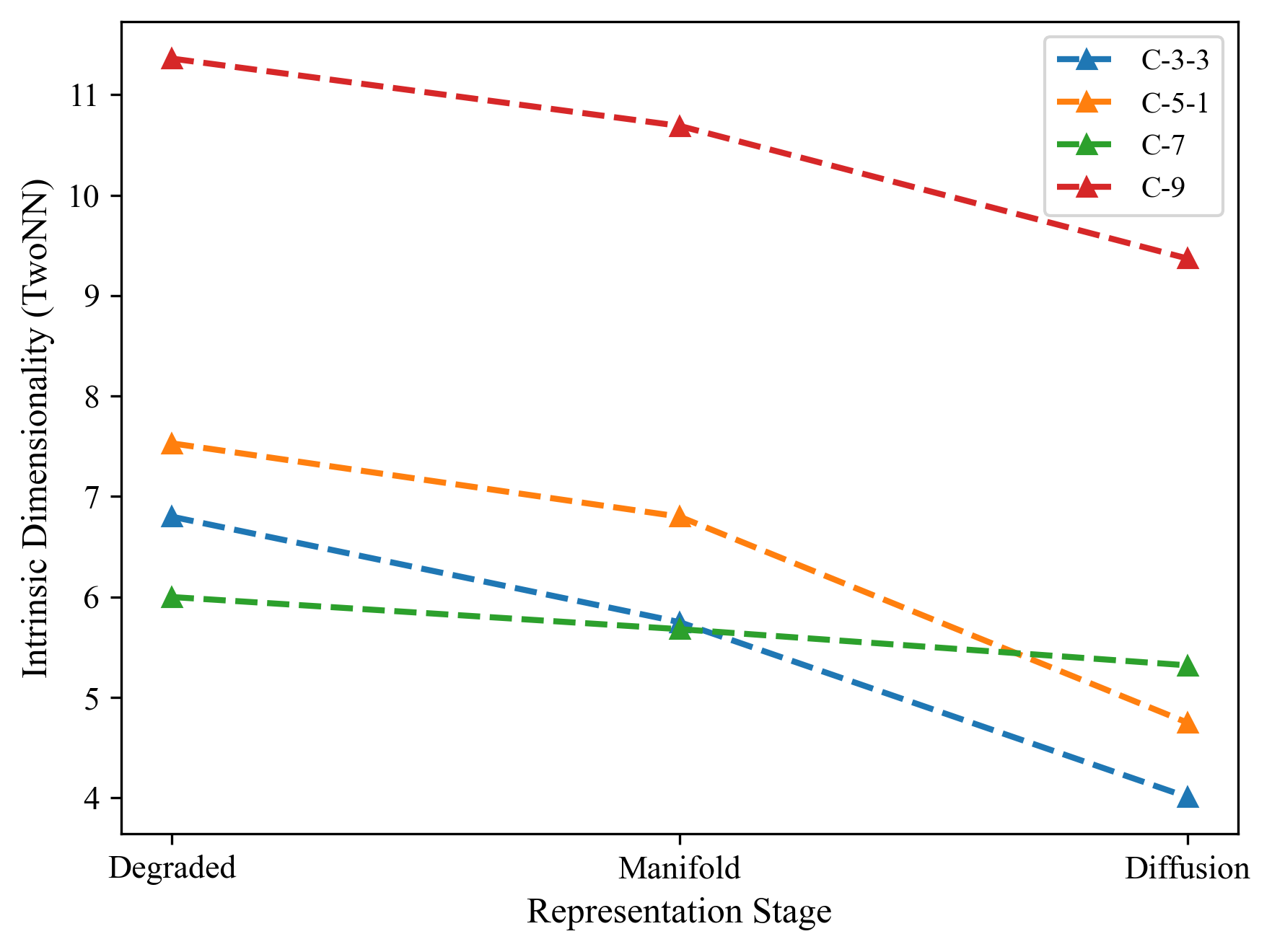} 
	\caption{Intrinsic dimensionality variation across representation stages under different composite degradation levels.} 
	\label{fig:line} 
\end{figure}

To complement the qualitative observations, Fig.~\ref{fig:line} reports the intrinsic dimensionality (ID) estimated by the TwoNN method across different representation stages. For all composite degradation levels, the intrinsic dimensionality consistently decreases from the degraded representation to the manifold embedding, and further to the diffusion-refined representation. This monotonic reduction quantitatively confirms that the proposed framework effectively suppresses degradation-induced high-dimensional perturbations and progressively restores a compact low-dimensional manifold structure, which provides a favorable geometric foundation for robust HSI classification.
\subsection{Efficiency Analysis}

\begin{table*}[htbp]
\centering
\small
\caption{Model Complexity Evaluation on PU and WHLK Datasets.}
\label{tab:params}
\begin{adjustbox}{width=0.8\textwidth}
\begin{tabular}{
  c|c|
  C{1.05cm}C{1.05cm}C{1.4cm}C{1.05cm}C{1.4cm}
  C{1.05cm}C{1.05cm}C{1.05cm}C{1.05cm}
}
\toprule
\textbf{Dataset} & \textbf{Metric} & \textbf{\footnotesize HyperTTA} & \textbf{\footnotesize SSEFN} & \textbf{\footnotesize SpectralDiff} & \textbf{\footnotesize SCLGC} & \textbf{\footnotesize RPCL-FSL} & \textbf{\footnotesize SS-MTr} & \textbf{\footnotesize DSFormer} & \textbf{\footnotesize Ours} \\
\midrule\midrule

\multirow{4}{*}{PU}
& Params (M)          
 & 0.761
 & 12.841
 & 4.305
 & 0.048
 & 0.072
 & 0.725
 & 0.677
 & 0.854\\

 & FLOPs (G)          
 & 0.207
 & 6.736
 & 8.096
 & 0.224
 & 0.174
 & 0.033
 & 0.060
 & 0.247 \\

 & Time (ms) 
 & 1.204
 & 6.854
 & 62.024
 & 0.187
 & 0.640
 & 1.291
 & 6.827
 & 5.483 \\
 
 & GPU Mem (GB) 
 & 0.066
 & 0.071
 & 0.254
 & 0.024
 & 0.011
 & 0.012
 & 0.014
 & 0.034 \\
\midrule

\multirow{4}{*}{WHLK}
  &Params (M)         
 & 1.217 & 18.826 & 6.673
 & 0.031 & 0.089 & 0.768
 & 0.677 & 0.854  \\
 & FLOPs (G) 
 & 0.525 & 9.888 & 10.801
 & 0.567 & 0.180 & 0.040
 & 0.060 & 0.260 \\
 & Time(ms)
 & 1.657 & 6.406 & 66.385
 & 0.303 & 0.646 & 1.311
 & 6.769 & 5.802 \\
 & GPU Mem(GB) 
 & 0.054 & 0.298 & 0.466
 & 0.063 & 0.011 & 0.012
 & 0.014 & 0.040 \\
 
\bottomrule
\end{tabular}
\end{adjustbox}
\end{table*}

\subsubsection{Manifold Anaysis}

The computational efficiency and practical feasibility of the proposed method are evaluated through comprehensive analyses of model complexity and runtime performance, including the number of parameters, FLOPs, inference time, and GPU memory consumption. All comparisons are conducted on standard benchmark datasets under identical evaluation settings, and the detailed results are summarized in Table \ref{tab:params}.

As summarized in Table \ref{tab:params}, MSDiff exhibits a moderate model size with 0.854 M parameters—higher than ultra-lightweight architectures like SCLGC and RPCL-FSL but substantially more compact than heavyweight models including SSEFN and SpectralDiff—thus striking a favorable balance between representational capacity and computational burden, which is critical for modeling complex composite degradation patterns without relying on excessively large networks; in terms of efficiency, it achieves a desirable trade-off with approximately 0.25 G FLOPs and an inference time of 5–6 ms per sample, being slightly costlier than ultra-lightweight baselines but significantly more efficient than heavy diffusion models such as SpectralDiff, thanks to performing diffusion in a compact low-dimensional manifold rather than the high-dimensional observation space, and furthermore, its modest GPU memory consumption (0.034–0.040 GB) makes it well-suited for practical deployment compared to memory-intensive alternatives like SSEFN.

Overall, the efficiency analysis demonstrates that MSDiff strikes a well-controlled and practically meaningful trade-off between computational cost and classification robustness. By confining diffusion modeling to a compact low-dimensional manifold rather than the original high-dimensional spectral–spatial space, the proposed method achieves strong robustness under complex degradation conditions while maintaining acceptable computational and memory overhead, making it suitable for real-world HSI analysis applications.

\section{Conclusion}

In this study, we proposed MSDiff to address the long-standing challenge of robust HSI classification under complex and composite degradation conditions. Motivated by the intrinsic high-dimensional but low-rank nature of hyperspectral data, the proposed method explicitly embeds degradation-corrupted observations into a compact low-dimensional spectral–spatial manifold and performs diffusion-based generative modeling directly within this manifold space. By jointly leveraging discriminative manifold embedding and manifold-aware diffusion refinement, MSDiff effectively suppresses degradation-induced high-dimensional disturbances while preserving intrinsic class-discriminative structures. Extensive experiments conducted on multiple hyperspectral benchmarks under diverse single and composite degradation settings demonstrate that MSDiff consistently outperforms state-of-the-art methods in terms of classification accuracy, stability, and robustness, validating the effectiveness of manifold-space generative modeling for degraded HSI classification. 

Despite its strong performance, the current framework is constrained by limitations in terms of patch-based scalability, the inherent rigidity of fixed diffusion schedules when modeling non-stationary degradations, and the absence of explicit full-scene spatial consistency. To mitigate these challenges, future research will focus on developing adaptive data-driven diffusion mechanisms for capturing evolving degradation patterns, integrating with large-scale foundation models to bolster cross-sensor generalization capabilities, and extending the model architecture toward end-to-end full-scene inference as well as broader remote sensing applications including spectral unmixing and land cover change detection.


%

\FloatBarrier

\ifCLASSOPTIONcaptionsoff
  \newpage
\fi



%
\bibliographystyle{IEEEtran}
\bibliography{HSI_References_Abbreviation}

\begin{thebibliography}{10}
\providecommand{\url}[1]{#1}
\csname url@samestyle\endcsname
\providecommand{\newblock}{\relax}
\providecommand{\bibinfo}[2]{#2}
\providecommand{\BIBentrySTDinterwordspacing}{\spaceskip=0pt\relax}
\providecommand{\BIBentryALTinterwordstretchfactor}{4}
\providecommand{\BIBentryALTinterwordspacing}{\spaceskip=\fontdimen2\font plus
\BIBentryALTinterwordstretchfactor\fontdimen3\font minus \fontdimen4\font\relax}
\providecommand{\BIBforeignlanguage}[2]{{%
\expandafter\ifx\csname l@#1\endcsname\relax
\typeout{** WARNING: IEEEtran.bst: No hyphenation pattern has been}%
\typeout{** loaded for the language `#1'. Using the pattern for}%
\typeout{** the default language instead.}%
\else
\language=\csname l@#1\endcsname
\fi
#2}}
\providecommand{\BIBdecl}{\relax}
\BIBdecl

\bibitem{10815625}
Z.~Zhang, L.~Huang, Q.~Wang, L.~Jiang, Y.~Qi, S.~Wang, T.~Shen, B.-H. Tang, and Y.~Gu, ``Uav hyperspectral remote sensing image classification: A systematic review,'' \emph{IEEE J. Sel. Top. Appl. Earth Obs. Remote Sens.}, vol.~18, pp. 3099--3124, 2025.

\bibitem{RAM2024109037}
\BIBentryALTinterwordspacing
B.~G. Ram, P.~Oduor, C.~Igathinathane, K.~Howatt, and X.~Sun, ``A systematic review of hyperspectral imaging in precision agriculture: Analysis of its current state and future prospects,'' \emph{Comput. Electron. Agric.}, vol. 222, p. 109037, 2024. [Online]. Available: \url{https://www.sciencedirect.com/science/article/pii/S0168169924004289}
\BIBentrySTDinterwordspacing

\bibitem{ADEP2017106}
\BIBentryALTinterwordspacing
R.~N. Adep, A.~shetty, and H.~Ramesh, ``Exhype: A tool for mineral classification using hyperspectral data,'' \emph{ISPRS J. Photogramm. Remote Sens.}, vol. 124, pp. 106--118, 2017. [Online]. Available: \url{https://www.sciencedirect.com/science/article/pii/S0924271616306645}
\BIBentrySTDinterwordspacing

\bibitem{DELUCA2024112}
\BIBentryALTinterwordspacing
G.~{De Luca}, F.~Carotenuto, L.~Genesio, M.~Pepe, P.~Toscano, M.~Boschetti, F.~Miglietta, and B.~Gioli, ``Improving prisma hyperspectral spatial resolution and geolocation by using sentinel-2: development and test of an operational procedure in urban and rural areas,'' \emph{ISPRS J. Photogramm. Remote Sens.}, vol. 215, pp. 112--135, 2024. [Online]. Available: \url{https://www.sciencedirect.com/science/article/pii/S0924271624002648}
\BIBentrySTDinterwordspacing

\bibitem{11322757}
N.~Li, Y.~Zhang, K.~Shi, Y.~Zhang, W.~Lin, X.~Luo, B.~Qin, G.~Zhu, and H.~Duan, ``Fusing hyperspectral proximal sensing to enhance chlorophyll-a estimation from sentinel-2,'' \emph{IEEE Trans. Geosci. Remote Sens.}, vol.~64, pp. 1--15, 2026.

\bibitem{11045957}
T.~Zhan, J.~Qi, J.~Zhang, X.~Yu, Q.~Du, and Z.~Wu, ``Spatial–spectral feature-enhanced mamba and sam-guided hyperspectral multiclass change detection,'' \emph{IEEE Trans. Geosci. Remote Sens.}, vol.~63, pp. 1--13, 2025.

\bibitem{SONG2025103285}
\BIBentryALTinterwordspacing
Y.~Song, J.~Zhang, Z.~Liu, Y.~Xu, S.~Quan, L.~Sun, J.~Bi, and X.~Wang, ``Deep learning for hyperspectral image classification: A comprehensive review and future predictions,'' \emph{Inf. Fusion}, vol. 123, p. 103285, 2025. [Online]. Available: \url{https://www.sciencedirect.com/science/article/pii/S1566253525003586}
\BIBentrySTDinterwordspacing

\bibitem{9552462}
L.~Zhuang and M.~K. Ng, ``Fasthymix: Fast and parameter-free hyperspectral image mixed noise removal,'' \emph{IEEE Trans. Neural Networks Learn. Syst.}, vol.~34, no.~8, pp. 4702--4716, 2023.

\bibitem{10637422}
X.~Wang, J.~Liu, Y.~Ni, W.~Chi, and Y.~Fu, ``Two-stage domain alignment single-source domain generalization network for cross-scene hyperspectral images classification,'' \emph{IEEE Trans. Geosci. Remote Sens.}, vol.~62, pp. 1--14, 2024.

\bibitem{10645292}
J.~Ye, F.~Xiong, J.~Zhou, and Y.~Qian, ``Iterative low-rank network for hyperspectral image denoising,'' \emph{IEEE Trans. Geosci. Remote Sens.}, vol.~62, pp. 1--15, 2024.

\bibitem{8359412}
S.~Li, R.~Dian, L.~Fang, and J.~M. Bioucas-Dias, ``Fusing hyperspectral and multispectral images via coupled sparse tensor factorization,'' \emph{IEEE Trans. Image Process.}, vol.~27, no.~8, pp. 4118--4130, 2018.

\bibitem{8603806}
R.~Dian, S.~Li, and L.~Fang, ``Learning a low tensor-train rank representation for hyperspectral image super-resolution,'' \emph{IEEE Trans. Neural Networks Learn. Syst.}, vol.~30, no.~9, pp. 2672--2683, 2019.

\bibitem{8677267}
H.~Huang, G.~Shi, H.~He, Y.~Duan, and F.~Luo, ``Dimensionality reduction of hyperspectral imagery based on spatial–spectral manifold learning,'' \emph{IEEE Trans. Cybern.}, vol.~50, no.~6, pp. 2604--2616, 2020.

\bibitem{10890865}
X.-H. Han and J.~Wang, ``Multi-degradation oriented deep unfolding model for hyperspectral image reconstruction,'' in \emph{Proc. IEEE ICASSP}, 2025, pp. 1--5.

\bibitem{li2025back}
\BIBentryALTinterwordspacing
T.~Li and K.~He, ``Back to basics: Let denoising generative models denoise,'' \emph{arXiv:2511.13720}, 2025. [Online]. Available: \url{https://arxiv.org/abs/2511.13720}
\BIBentrySTDinterwordspacing

\bibitem{2025Efficient}
M.~Zhang, Z.~Lei, L.~Liu, K.~Ma, R.~Shang, and L.~Jiao, ``Efficient evolutionary multi-scale spectral-spatial attention fusion network for hyperspectral image classification,'' \emph{Expert Syst. Appl.}, no. Mar., p. 262, 2025.

\bibitem{10843260}
B.~Xi, Y.~Zhang, J.~Li, T.~Zheng, X.~Zhao, H.~Xu, C.~Xue, Y.~Li, and J.~Chanussot, ``Mctgcl: Mixed cnn–transformer for mars hyperspectral image classification with graph contrastive learning,'' \emph{IEEE Trans. Geosci. Remote Sens.}, vol.~63, pp. 1--14, 2025.

\bibitem{rs14091968}
\BIBentryALTinterwordspacing
J.~Sun, J.~Zhang, X.~Gao, M.~Wang, D.~Ou, X.~Wu, and D.~Zhang, ``Fusing spatial attention with spectral-channel attention mechanism for hyperspectral image classification via encoder–decoder networks,'' \emph{Remote Sens.}, vol.~14, no.~9, 2022. [Online]. Available: \url{https://www.mdpi.com/2072-4292/14/9/1968}
\BIBentrySTDinterwordspacing

\bibitem{9684381}
L.~Sun, G.~Zhao, Y.~Zheng, and Z.~Wu, ``Spectral–spatial feature tokenization transformer for hyperspectral image classification,'' \emph{IEEE Trans. Geosci. Remote Sens.}, vol.~60, pp. 1--14, 2022.

\bibitem{ahmad2025comprehensive}
M.~Ahmad, S.~Distefano, A.~M. Khan, M.~Mazzara, C.~Li, H.~Li, J.~Aryal, Y.~Ding, G.~Vivone, and D.~Hong, ``A comprehensive survey for hyperspectral image classification: The evolution from conventional to transformers and mamba models,'' \emph{Neurocomputing}, p. 130428, 2025.

\bibitem{10976442}
Y.~He, B.~Tu, B.~Liu, J.~Li, and A.~Plaza, ``Hsi-mformer: Integrating mamba and transformer experts for hyperspectral image classification,'' \emph{IEEE Trans. Geosci. Remote Sens.}, vol.~63, pp. 1--16, 2025.

\bibitem{rs14205199}
\BIBentryALTinterwordspacing
C.~Li, Z.~Li, X.~Liu, and S.~Li, ``The influence of image degradation on hyperspectral image classification,'' \emph{Remote Sens.}, vol.~14, no.~20, 2022. [Online]. Available: \url{https://www.mdpi.com/2072-4292/14/20/5199}
\BIBentrySTDinterwordspacing

\bibitem{8877854}
C.~Wang, L.~Zhang, W.~Wei, and Y.~Zhang, ``Hyperspectral image classification with data augmentation and classifier fusion,'' \emph{IEEE Geosci. Remote Sens. Lett.}, vol.~17, no.~8, pp. 1420--1424, 2020.

\bibitem{9439796}
T.~Li and Y.~Gu, ``Progressive spatial–spectral joint network for hyperspectral image reconstruction,'' \emph{IEEE Trans. Geosci. Remote Sens.}, vol.~60, pp. 1--14, 2022.

\bibitem{10445289}
S.~Hu, F.~Gao, X.~Zhou, J.~Dong, and Q.~Du, ``Hybrid convolutional and attention network for hyperspectral image denoising,'' \emph{IEEE Geosci. Remote Sens. Lett.}, vol.~21, pp. 1--5, 2024.

\bibitem{8435923}
Y.~Chang, L.~Yan, H.~Fang, S.~Zhong, and W.~Liao, ``Hsi-denet: Hyperspectral image restoration via convolutional neural network,'' \emph{IEEE Trans. Geosci. Remote Sens.}, vol.~57, no.~2, pp. 667--682, 2019.

\bibitem{9376953}
T.~Li, Y.~Cai, Z.~Cai, X.~Liu, and Q.~Hu, ``Nonlocal band attention network for hyperspectral image band selection,'' \emph{IEEE J. Sel. Top. Appl. Earth Obs. Remote Sens.}, vol.~14, pp. 3462--3474, 2021.

\bibitem{9125995}
R.~Hang, F.~Zhou, Q.~Liu, and P.~Ghamisi, ``Classification of hyperspectral images via multitask generative adversarial networks,'' \emph{IEEE Trans. Geosci. Remote Sens.}, vol.~59, no.~2, pp. 1424--1436, 2021.

\bibitem{hypertta}
\BIBentryALTinterwordspacing
X.~Yue, A.~Liu, N.~Chen, C.~Huang, H.~Liu, Z.~Huang, and L.~Fang, ``Hypertta: Test-time adaptation for hyperspectral image classification under distribution shifts,'' \emph{arXiv:2509.08436}, 2025. [Online]. Available: \url{https://arxiv.org/abs/2509.08436}
\BIBentrySTDinterwordspacing

\bibitem{8542643}
W.~Li, C.~Chen, M.~Zhang, H.~Li, and Q.~Du, ``Data augmentation for hyperspectral image classification with deep cnn,'' \emph{IEEE Geosci. Remote Sens. Lett.}, vol.~16, no.~4, pp. 593--597, 2019.

\bibitem{9408225}
Z.~Gong, W.~Hu, X.~Du, P.~Zhong, and P.~Hu, ``Deep manifold embedding for hyperspectral image classification,'' \emph{IEEE Trans. Cybern.}, vol.~52, no.~10, pp. 10\,430--10\,443, 2022.

\bibitem{chen2025spectral}
\BIBentryALTinterwordspacing
W.~Chen, Y.~Zhang, Z.~Xiao, J.~Chu, and X.~Wang, ``Spectral-spatial self-supervised learning for few-shot hyperspectral image classification,'' \emph{arXiv:2505.12482}, 2025. [Online]. Available: \url{https://arxiv.org/abs/2505.12482}
\BIBentrySTDinterwordspacing

\bibitem{10193882}
Z.~Li, H.~Guo, Y.~Chen, C.~Liu, Q.~Du, Z.~Fang, and Y.~Wang, ``Few-shot hyperspectral image classification with self-supervised learning,'' \emph{IEEE Trans. Geosci. Remote Sens.}, vol.~61, pp. 1--17, 2023.

\bibitem{11050958}
J.~Dong, M.~Liang, Z.~He, and C.~Zhou, ``Hierarchical and bidirectional contrastive learning for hyperspectral image classification,'' \emph{IEEE Trans. Geosci. Remote Sens.}, vol.~63, pp. 1--17, 2025.

\bibitem{zhang2025graph}
Y.~Zhang, Q.~Han, L.~Wang, K.~Cheng, B.~Wang, and K.~Zhan, ``Graph-weighted contrastive learning for semi-supervised hyperspectral image classification,'' \emph{J. Electron. Imaging}, vol.~34, no.~2, pp. 023\,044--023\,044, 2025.

\bibitem{10459061}
N.~A.~A. Braham, J.~Mairal, J.~Chanussot, L.~Mou, and X.~X. Zhu, ``Enhancing contrastive learning with positive pair mining for few-shot hyperspectral image classification,'' \emph{IEEE J. Sel. Top. Appl. Earth Obs. Remote Sens.}, vol.~17, pp. 8509--8526, 2024.

\bibitem{Spectraldiff}
N.~Chen, J.~Yue, L.~Fang, and S.~Xia, ``Spectraldiff: A generative framework for hyperspectral image classification with diffusion models,'' \emph{IEEE Trans. Geosci. Remote Sens.}, vol.~61, pp. 1--16, 2023.

\bibitem{zhang2024data}
J.~Zhang, F.~Zhao, H.~Liu, and J.~Yu, ``Data and knowledge-driven deep multiview fusion network based on diffusion model for hyperspectral image classification,'' \emph{Expert Syst. Appl.}, vol. 249, p. 123796, 2024.

\bibitem{lu2025beyond}
M.~Lu, N.~Chen, X.~Yue, Y.~Li, J.~Yue, and L.~Fang, ``Beyond dimensionality explosion: A latent diffusion framework for hyperspectral image classification,'' \emph{Neurocomputing}, p. 131249, 2025.

\bibitem{qu2024mtlsc}
J.~Qu, L.~Xiao, W.~Dong, and Y.~Li, ``Mtlsc-diff: Multitask learning with diffusion models for hyperspectral image super-resolution and classification,'' \emph{Knowledge-Based Syst.}, vol. 303, p. 112415, 2024.

\bibitem{hu2025}
\BIBentryALTinterwordspacing
X.~Hu, X.~Liu, D.~Hong, Q.~Duan, L.~Jiang, H.~Yang, and D.~Zhan, ``Recent advances in diffusion models for hyperspectral image processing and analysis: A review,'' \emph{arXiv:2505.11158}, 2025. [Online]. Available: \url{https://arxiv.org/abs/2505.11158}
\BIBentrySTDinterwordspacing

\bibitem{8573977}
Y.~Zhong, X.~Wang, Y.~Xu, S.~Wang, T.~Jia, X.~Hu, J.~Zhao, L.~Wei, and L.~Zhang, ``Mini-uav-borne hyperspectral remote sensing: From observation and processing to applications,'' \emph{IEEE Geosci. Remote Sens. Mag.}, vol.~6, no.~4, pp. 46--62, 2018.

\bibitem{ZHONG2020112012}
Y.~Zhong, X.~Hu, C.~Luo, X.~Wang, J.~Zhao, and L.~Zhang, ``Whu-hi: Uav-borne hyperspectral with high spatial resolution (h2) benchmark datasets and classifier for precise crop identification based on deep convolutional neural network with crf,'' \emph{Remote Sens. Environ.}, vol. 250, p. 112012, 2020.

\bibitem{ssmtr}
L.~Huang, Y.~Chen, and X.~He, ``Spectral–spatial masked transformer with supervised and contrastive learning for hyperspectral image classification,'' \emph{IEEE Trans. Geosci. Remote Sens.}, vol.~61, pp. 1--18, 2023.

\bibitem{DSFormer}
\BIBentryALTinterwordspacing
Y.~Xu, D.~Wang, L.~Zhang, and L.~Zhang, ``Dual selective fusion transformer network for hyperspectral image classification,'' \emph{Neural Networks}, vol. 187, p. 107311, 2025. [Online]. Available: \url{https://www.sciencedirect.com/science/article/pii/S089360802500190X}
\BIBentrySTDinterwordspacing

\bibitem{slcgc}
Y.~Ding, Z.~Zhang, A.~Yang, Y.~Cai, X.~Xiao, D.~Hong, and J.~Yuan, ``Slcgc: A lightweight self-supervised low-pass contrastive graph clustering network for hyperspectral images,'' \emph{IEEE Trans. Multimedia}, vol.~27, pp. 8251--8262, 2025.

\bibitem{RPCL}
Q.~Liu, J.~Peng, Y.~Ning, N.~Chen, W.~Sun, Q.~Du, and Y.~Zhou, ``Refined prototypical contrastive learning for few-shot hyperspectral image classification,'' \emph{IEEE Trans. Geosci. Remote Sens.}, vol.~61, pp. 1--14, 2023.

\bibitem{SSEFN}
S.~Liu, C.~Fu, Y.~Duan, X.~Wang, and F.~Luo, ``Spatial–spectral enhancement and fusion network for hyperspectral image classification with few labeled samples,'' \emph{IEEE Trans. Geosci. Remote Sens.}, vol.~63, pp. 1--14, 2025.

\end{thebibliography}

\FloatBarrier
\end{document}